\documentclass[acmsmall]{acmart}

\usepackage{multirow}
\usepackage{amsfonts}
\usepackage{color}
\usepackage{subfigure}
\usepackage{adjustbox,lipsum}
\usepackage{array}

\definecolor{orange}{rgb}{1,0.5,0}
\definecolor{grey}{rgb}{0.5,0.5,0.5}
\definecolor{new_blue}{rgb}{0.3,0.5,0.8}

\newcommand{\PreserveBackslash}[1]{\let\temp=\\#1\let\\=\temp}
\newcolumntype{C}[1]{>{\PreserveBackslash\centering}p{#1}}
\newcolumntype{R}[1]{>{\PreserveBackslash\raggedleft}p{#1}}
\newcolumntype{L}[1]{>{\PreserveBackslash\raggedright}p{#1}}

\AtBeginDocument{%
  \providecommand\BibTeX{{%
    \normalfont B\kern-0.5em{\scshape i\kern-0.25em b}\kern-0.8em\TeX}}}

\setcopyright{acmcopyright}
\acmJournal{TOMM}
\acmYear{2020} \acmVolume{1} \acmNumber{1} \acmArticle{1} \acmMonth{1} \acmPrice{15.00}\acmDOI{10.1145/3419439}

\begin{document}

\title{GreyReID: A Novel Two-stream Deep Framework with RGB-grey Information for Person Re-identification}

\author{Lei Qi}
\affiliation{%
  \institution{The State Key Laboratory for Novel Software Technology, Nanjing University}
  \city{Nanjing}
  \country{China}}
\email{qilei.cs@gmail.com}

\author{Lei Wang}
\affiliation{%
  \institution{School of Computing and Information Technology, University of Wollongong}
  \city{Wollongong}
  \country{Australia}
  }\email{leiw@uow.edu.au}

\author{Jing Huo}
\affiliation{%
 \institution{The State Key Laboratory for Novel Software Technology, Nanjing University}
 \city{Nanjing}
 \country{China}
 }
 \email{huojing@nju.edu.cn}


\author{Yinghuan Shi}
\affiliation{%
  \institution{The State Key Laboratory for Novel Software Technology, Nanjing University}
 \city{Nanjing}
 \country{China}
 }\email{syh@nju.edu.cn}

\author{Yang Gao}
\affiliation{\institution{The State Key Laboratory for Novel Software Technology, Nanjing University}
 \city{Nanjing}
 \country{China}
 }\email{gaoy@nju.edu.cn}

\thanks{This work was supported by National Key Research and Development Program of China (No. 2019YFC0118300), Science and Technology Innovation 2030-"New Generation Artificial Intelligence" Major Project (No. 2018AAA0100900 and 2018AAA0100905), Natural Science Foundation of China (No. 61806092 and 61673203) and Jiangsu Natural Science Foundation (No. BK20180326). (Corresponding author: Yang Gao.)}


\begin{abstract}
In this paper, we observe that most false positive images (i.e., different identities with query images) in the top ranking list usually have the similar color information with the query image in person re-identification (Re-ID). Meanwhile, when we use the greyscale images generated from RGB images to conduct the person Re-ID task, some hard query images can obtain better performance compared with using RGB images. Therefore, RGB and greyscale images seem to be complementary to each other for person Re-ID. In this paper, we aim to utilize both RGB and greyscale images to improve the person Re-ID performance. To this end, we propose a novel two-stream deep neural network with RGB-grey information, which can effectively fuse RGB and greyscale feature representations to enhance the generalization ability of Re-ID. Firstly, we convert RGB images to greyscale images in each training batch. Based on these RGB and greyscale images, we train the RGB and greyscale branches, respectively. Secondly, to build up connections between RGB and greyscale branches, we merge the RGB and greyscale branches into a new joint branch. Finally, we concatenate the features of all three branches as the final feature representation for Re-ID. Moreover, in the training process, we adopt the joint learning scheme to simultaneously train each branch by the independent loss function, which can enhance the generalization ability of each branch. Besides, a global loss function is utilized to further fine-tune the final concatenated feature. The extensive experiments on multiple benchmark datasets fully show that the proposed method can outperform the state-of-the-art person Re-ID methods. Furthermore, using greyscale images can indeed improve the person Re-ID performance in the proposed deep framework.
\end{abstract}




\maketitle
\section{Introduction}
Person re-identification (Re-ID) aims at finding the interesting person from a large-scale gallery set which consists of many person images with different illumination, pose, resolution and background from many non-overlapping camera views~\cite{fan2018unsupervised,qi2018unsupervised,qi2019novel}. These discrepancies of different camera views result in challenges in person Re-ID. In addition, the similarity among many different person images is also a great challenge for person Re-ID. For example, many person images with different identities could have a similar appearance, especially for the color information. 

Many methods have been developed to deal with the above challenges~\cite{DBLP:conf/mm/LiuZXXZ18,DBLP:conf/mm/HanGZZ18,DBLP:conf/cvpr/TianYLLZSYW18,DBLP:conf/cvpr/KalayehBGKS18,DBLP:conf/cvpr/Song0O018,DBLP:conf/cvpr/Zhong0ZL018,DBLP:conf/cvpr/LiuNYZCH18,DBLP:conf/eccv/WangZHLW18,zheng2018re}. In recent years, using extra information, such as  person attribute~\cite{DBLP:conf/mm/LiuZXXZ18,DBLP:conf/mm/HanGZZ18,zhao2019attribute} and the image-segmentation information~\cite{DBLP:conf/cvpr/TianYLLZSYW18,DBLP:conf/cvpr/KalayehBGKS18,DBLP:conf/cvpr/Song0O018}, has significantly improved the Re-ID performance. Particularly, using image segmentation can effectively reduce  the impact of the cluttered background. Besides, to avoid the model over-fitting, some extra images are generated by GANs-based methods to enhance the diversity of training data~\cite{DBLP:conf/cvpr/Zhong0ZL018,DBLP:conf/cvpr/LiuNYZCH18,song2019unsupervised}. Furthermore, the local-based methods can effectively boost the Re-ID performance, which include the part-based methods~\cite{DBLP:conf/eccv/SunZYTW18,DBLP:conf/mm/WeiZY0T17,DBLP:conf/iccv/SuLZX0T17} and the attention-based methods~\cite{DBLP:conf/eccv/WangZHLW18,zheng2018re,liu2017end}. However, the above methods mainly focus on the challenges caused by the discrepancy of different camera views, which have not paid particular attention to the similarity of the color information among different identities.
\begin{figure*}
\centering
\subfigure[CUHK03]{
\includegraphics[width=6.7cm]{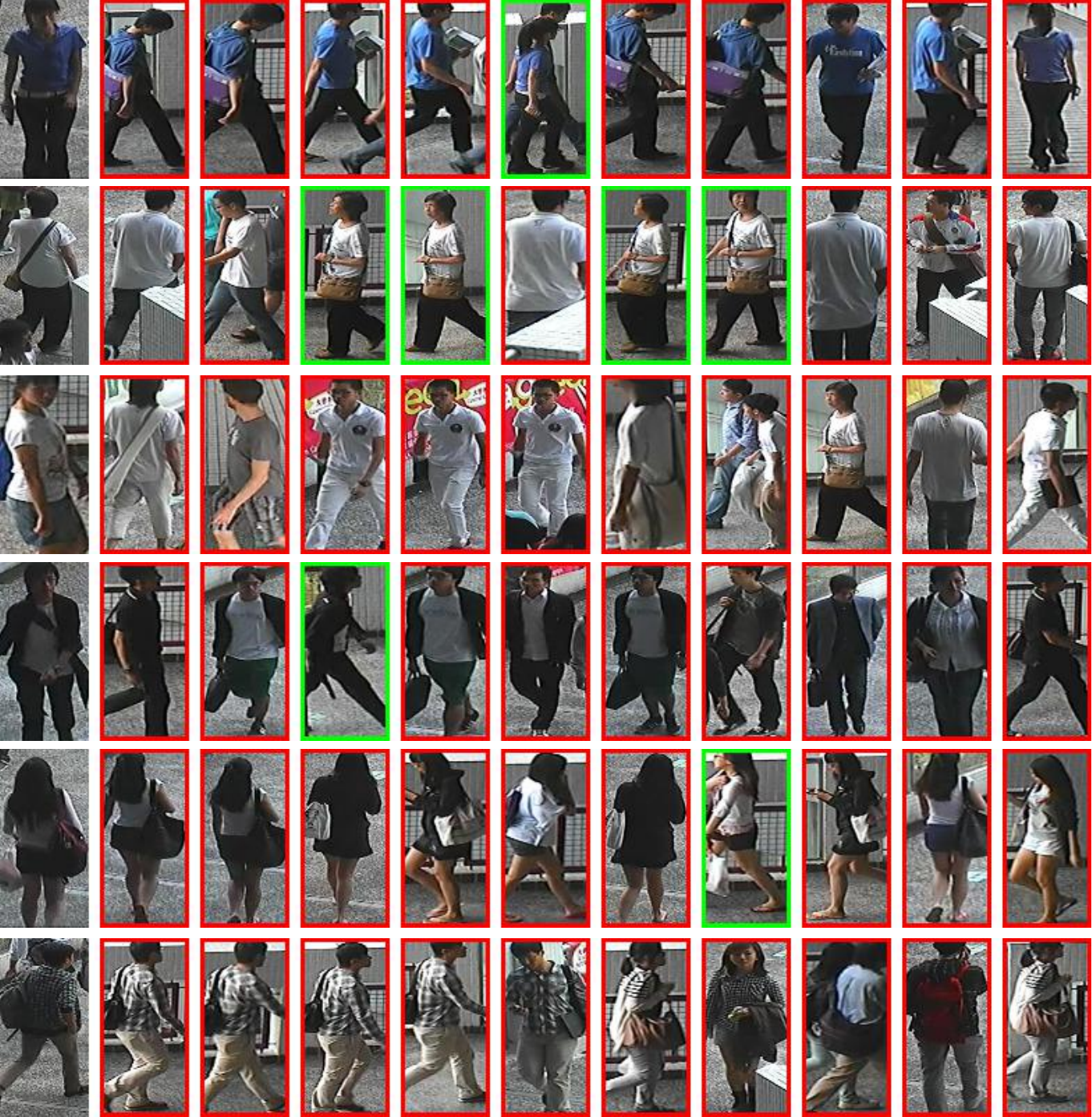}
}
\subfigure[Market1501]{
\includegraphics[width=6.7cm]{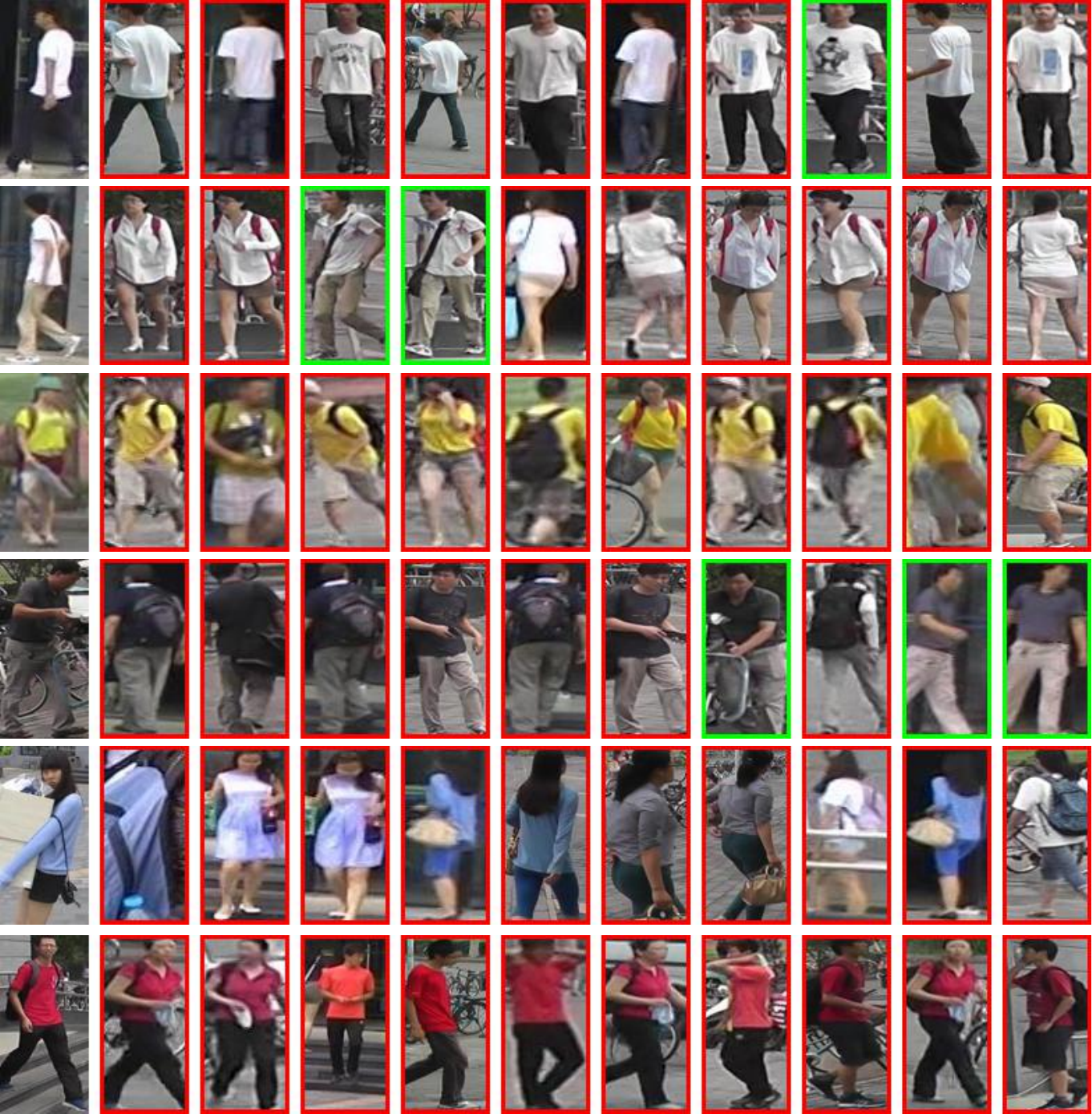}
}
\caption{Person retrieval samples of the PCB model~\cite{DBLP:conf/eccv/SunZYTW18} on CUHK03 and Market1501, respectively. Note that the first image of each row is the query image and other images are ``Rank-1 to Rank-10'' in the ranking list. The {\color{green}green} ({\color{red}red}) boxes denote the {\color{green}positive} ({\color{red}negative}) images with the query image. As seen in this figure, the negative samples in the top positions have similar color information with the query image.}
\label{fig1} 
\end{figure*}

In practical applications, there are many similar person images with different identities on the Re-ID dataset, especially in terms of the color information. Thus, for a query image, these similar images with different identities may appear at the top positions of the ranking list. For example, Fig.~\ref{fig1} visualizes the ranking list of the PCB model~\cite{DBLP:conf/eccv/SunZYTW18}, which has obtained the state-of-the-art results on two Re-ID datasets. As seen, for most false positive images in the top ranking list, they have similar color information with query images, which is called the color over-fitting of person Re-ID in this paper. Thus, many existing methods could overly depend on color information and neglect some structure information in person images.

\begin{figure*}
\centering
\subfigure[CUHK03]{
\includegraphics[width=6.7cm]{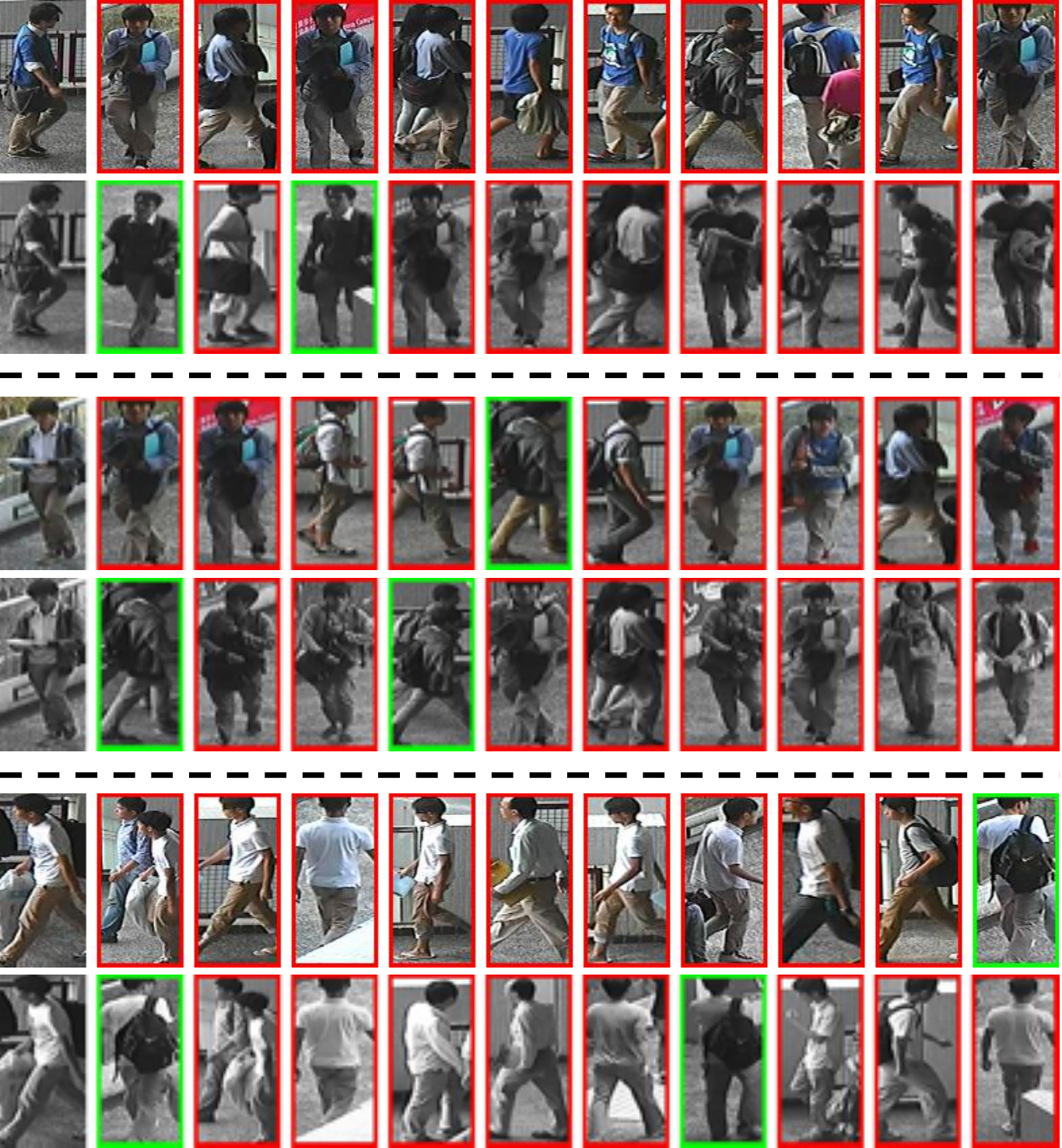}
}
\subfigure[Market1501]{
\includegraphics[width=6.7cm]{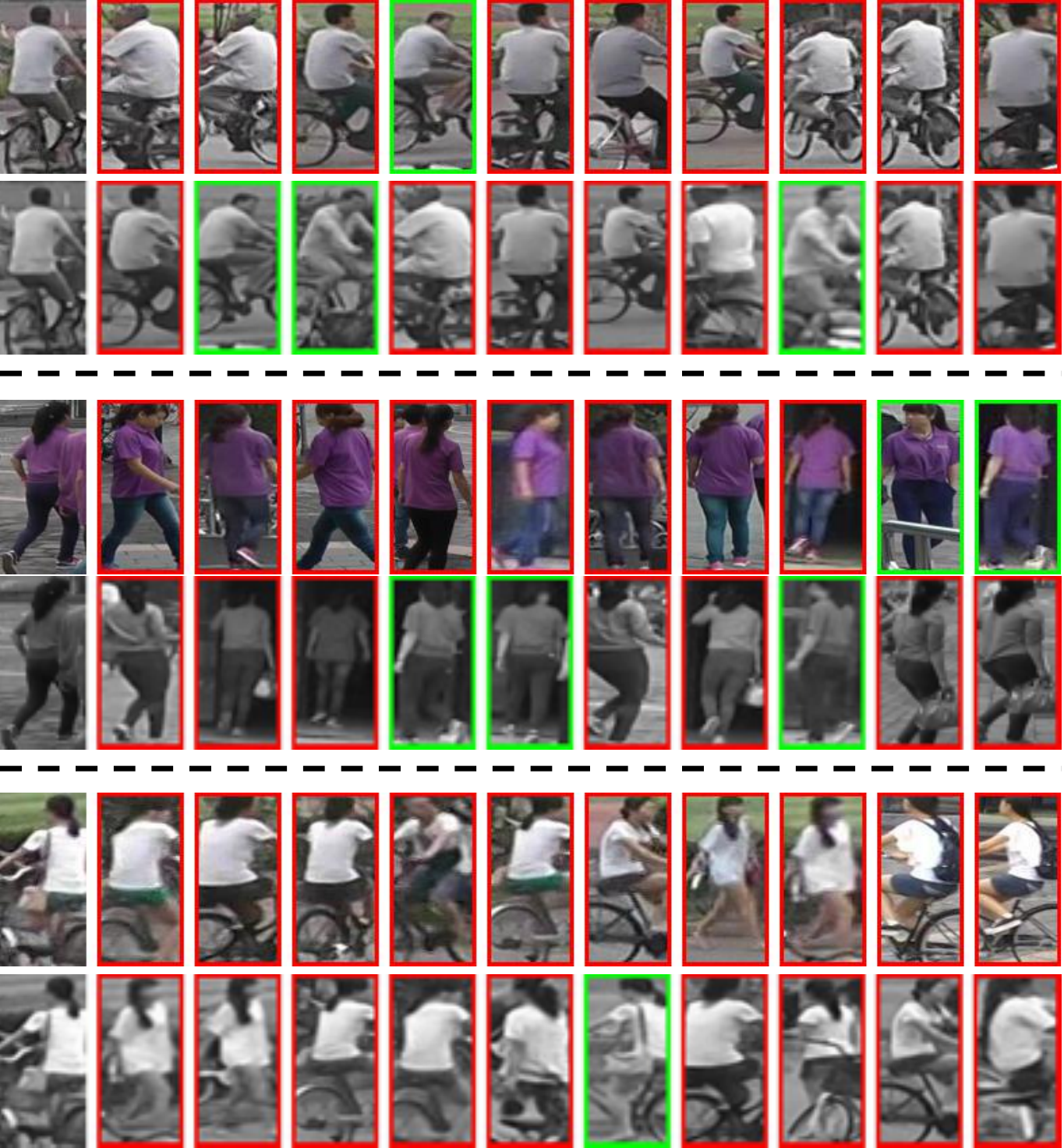}
}
\caption{Person retrieval samples of greyscale and RGB images in the PCB model~\cite{DBLP:conf/eccv/SunZYTW18} on CUHK03 and Market1501, respectively. In particular, PCB is trained on greyscale and RGB images, respectively. Top (bottom) of each pair is retrieval samples of RGB (greyscale) images. Note that the first image of each row is the query image and other images are ``Rank-1 to Rank-10'' in the ranking list. The {\color{green}green} ({\color{red}red}) boxes denote the {\color{green}positive} ({\color{red}negative}) images with the query image. As seen in this figure, there is the complementarity between RGB and greyscale images.}
\label{fig2}
\end{figure*}
To solve this issue, we utilize greyscale images to conduct the person Re-ID task. Compared with RGB images, greyscale images remove the potentially disturbing color information, and thus using them may make deep models better focus on other information besides the color information, such as structure and texture information. Through the experiments, we find that using greyscale images to conduct the person Re-ID task can generate the complementary results with RGB images, as shown in Fig.~\ref{fig2}. Some hard query images may fail by using features of RGB images, but utilizing greyscale images can achieve better results. Therefore, we argue that RGB and greyscale images have the complementarity in the person Re-ID task.However, since greyscale images could bring the noisy information into the training model, we do not simply use them based on the data-augmentation scheme, which cannot obtain better performance as reported in Table~\ref{tab04}. Therefore, we need to design a new deep framework to address this issue.

In this paper, to exploit the complementarity between RGB and greyscale images, we propose a two-stream deep framework with RGB-grey information for person Re-ID. Firstly, a two-stream network is designed with RGB-greyscale image pairs as inputs. In particular, greyscale images are generated by simply converting RGB images. Then, we fuse the RGB and greyscale branches into a joint branch, which can further extract a more robust feature than RGB and grey features alone. Finally, we concatenate the greyscale, RGB and joint features as the final person representation. Moreover, to further enhance the generalization ability of all three branches, an independent loss function is used for each branch in the training stage. In addition, we also employ a global loss to further optimize the final concatenated feature. The proposed method can adequately explore the complementary information between RGB and greyscale images to boost the person Re-ID performance.

In summary, we observe that greyscale information has an important role in person Re-ID and put forward an end-to-end deep framework to combine RGB and Greyscale information in person image. Our contribution in this work is threefold.
First, we find that RGB and greyscale images are complementary in person Re-ID. To the best of our survey, the proposed method is the first to consider the color over-fitting issue in person Re-ID and employs greyscale person images to mitigate this issue.
Second, to fully explore the complementary information in RGB and greyscale images, we develop a two-stream network with RGB and greyscale information for person Re-ID, and employ an independent loss function for each branch in the proposed framework.    
Last, extensive experiments on multiple benchmark datasets demonstrate the efficacy of greyscale images in the person Re-ID task. Furthermore, the proposed method can also obtain competitive performance when compared with the state-of-the-art methods.

The rest of this paper is organized as follows.
The related work is reviewed in Section \ref{s-related}.
The framework proposed in this work is elaborated and discussed in Section \ref{s-framework}.
Experimental results and analysis are presented in Section \ref{s-experiment},
and the conclusion is drawn in Section \ref{s-conclusion}.
\section{Related Work}\label{s-related}
In this section, we review the most related work with this paper in the person Re-ID community.
\subsection{Person Re-ID with extra information}
In person Re-ID, extra information, such as  person attribute~~\cite{DBLP:conf/mm/LiuZXXZ18,DBLP:conf/mm/HanGZZ18,tay2019aanet} and the image-segmentation information~\cite{DBLP:conf/cvpr/TianYLLZSYW18,DBLP:conf/cvpr/KalayehBGKS18,DBLP:conf/cvpr/Song0O018,qi2018maskreid}, can effectively help to improve the Re-ID performance.
Liu \textit{et al.}~\cite{DBLP:conf/mm/LiuZXXZ18} present a novel Contextual-attentional Attribute-appearance Network (CA$^3$NET) for person Re-ID, which simultaneously exploits the complementarity between semantic attributes and visual appearance.
In ~\cite{DBLP:conf/mm/HanGZZ18},  a novel Attribute-Aware Attention Model (A$^3$M) is developed, which can learn local attribute representation and global category representation simultaneously in an end-to-end manner. 
In addition, due to the diverse background clutters from different camera views, the segmentation information can alleviate background clutters for pedestrian images, and improve the performance of person Re-ID.
In ~\cite{DBLP:conf/cvpr/KalayehBGKS18}, the proposed method integrates human semantic parsing in person Re-ID and not only considerably outperforms its counter baseline, but also achieves state-of-the-art performance. Besides,
Song \textit{et al.}~\cite{DBLP:conf/cvpr/Song0O018} introduce the binary segmentation masks to construct synthetic RGB-Mask pairs as inputs, then they design a Mask-guided Contrastive Attention Model (MGCAM) to learn features separately from the body and background regions. Unlike the above methods, considering the color over-fitting issue in person Re-ID, we introduce the greyscale information to enhance the generalization ability of person Re-ID, which is trivially captured by converting RGB images.

\subsection{Person Re-ID with extra images}
To enrich training samples, some extra images can be generated by GANs-based methods~\cite{DBLP:conf/iccv/ZhengZY17,DBLP:conf/cvpr/Zhong0ZL018,DBLP:conf/cvpr/LiuNYZCH18,DBLP:conf/cvpr/WeiZ0018,zhong2018generalizing}, which can reduce the model over-fitting. Zheng \textit{et al.}~\cite{DBLP:conf/iccv/ZhengZY17} employ GAN~\cite{DBLP:journals/corr/RadfordMC15} to generate unlabeled extra images in the Re-ID task. To use them, a uniform label distribution is assigned to these generated images, which regularizes the supervised model and improves the baseline.
Considering the pose variation in different camera views, Liu \textit{et al.}~\cite{DBLP:conf/cvpr/LiuNYZCH18} propose a pose-transferrable person Re-ID framework which utilizes pose-transferred sample augmentations to enhance Re-ID model training. 
Besides, person Re-ID also suffers from image style variations caused by different cameras. Zhong \textit{et al.}~\cite{DBLP:conf/cvpr/Zhong0ZL018} implicitly address this problem by learning a camera-invariant descriptor subspace. In particular, the method explicitly considers this challenge by introducing camera style adaptation which can serve as a data augmentation approach that smooths the camera style disparities. Different from the above methods, we introduce the extra greyscale images by converting RGB images directly, which does not need to train extra models, and mitigate the model over-fitting issue from the image color-level view. 

\subsection{The local-based person Re-ID}
In recent year, the local-based methods can significantly improve the performance of person Re-ID, which include the part-based methods~\cite{DBLP:conf/iccv/ZhengZY17,DBLP:conf/eccv/SunZYTW18,DBLP:conf/mm/WeiZY0T17,DBLP:conf/iccv/SuLZX0T17} and the attention-based methods~\cite{DBLP:conf/eccv/WangZHLW18,zheng2018re,si2018dual}.
Due to the huge variance of human pose and the misalignment of detected human images, 
Wei \textit{et al.}~\cite{DBLP:conf/mm/WeiZY0T17} propose a Global-Local-Alignment Descriptor (GLAD) which explicitly leverages the local and global cues in human body to
generate a discriminative and robust representation.
Su \textit{et al.}~\cite{DBLP:conf/iccv/SuLZX0T17} develop a Pose-driven Deep Convolutional (PDC) model to learn
improved feature extraction and matching models from end to end, which employs the human part cues to alleviate the pose variations and learn robust feature representations from both the global image and different local parts.
Instead of using external cues, such as pose estimation,
Sun \textit{et al.}~\cite{DBLP:conf/eccv/SunZYTW18} propose a network named Part-based Convolutional Baseline (PCB) which outputs a convolutional descriptor consisting of several part-level features. 
Besides, the attention-based model can also locate the local regions.
Wang \textit{et al.}~\cite{DBLP:conf/eccv/WangZHLW18} present a novel deep network called Mancs which utilizes the attention mechanism for the person misalignment problem
and properly sampling for the ranking loss to obtain a more stable person
representation.
Zheng \textit{et al.}~\cite{zheng2018re}  propose the Consistent Attentive Siamese Network (CASN), which enforces attention consistency among images of the same identity. Compared with the above methods, the proposed method integrates the different attention regions of greyscale and RGB images to learn the robust feature representations for person Re-ID.

\section{The Proposed Method}\label{s-framework}
In this part, we first present the complementarity between RGB and greyscale images in Section~\ref{grey}. Then the proposed two-stream deep framework with RGB-grey information is described in Section~\ref{RGFFN}. Lastly, we further discuss some components of the proposed method in Section~\ref{Discussion}.
\subsection{The complementarity between RGB and greyscale images}~\label{grey}
\begin{figure}
\centering
\includegraphics[width=14cm]{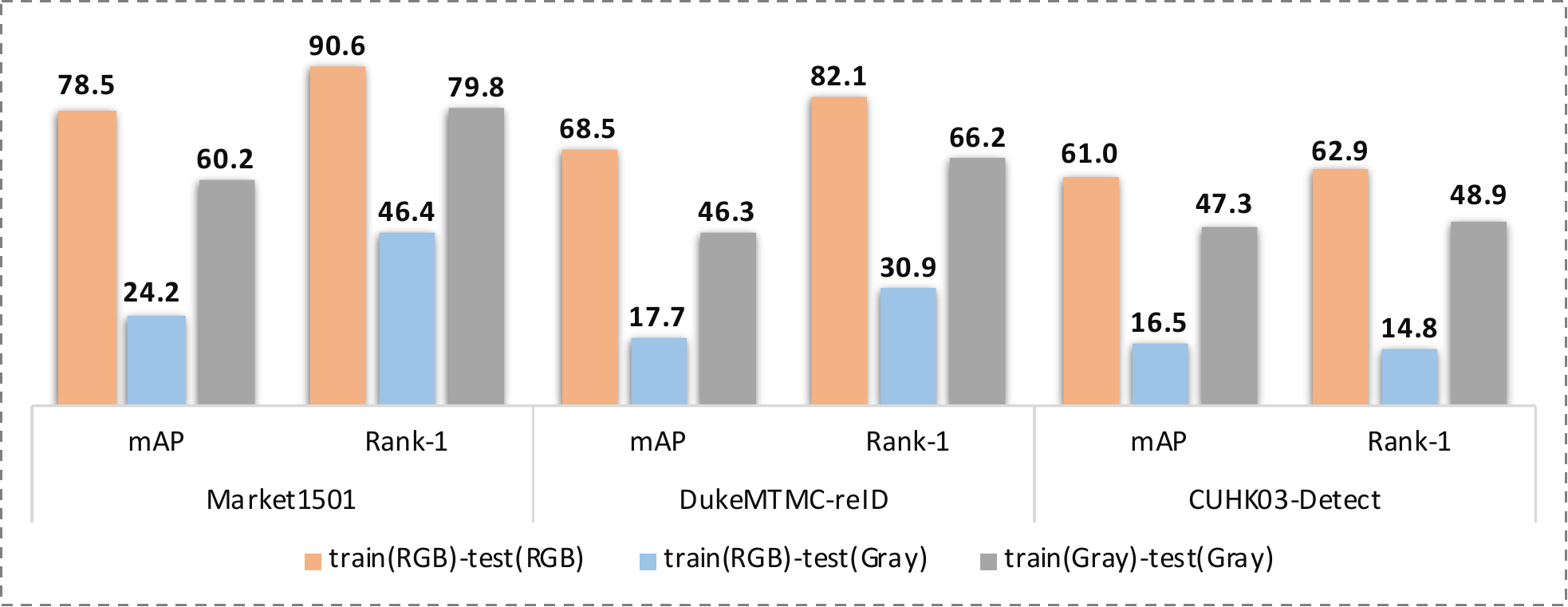}
\caption{The evaluation of the greyscale and RGB models on Market1501, DukeMTMC-reID and CUHK03, respectively. Note that {\color{orange}orange} ({\color{grey}grey}) bars denote using RGB (greyscale) images to train and test ResNet-50. {\color{new_blue}Blue} bars represent using RGB and greyscale images to train and test the ResNet-50 model, respectively.}
\label{fig7}
\end{figure}

In person Re-ID, most existing methods utilize RGB color images to train and test models. However, these methods may overly depend on the color information (i.e., the color over-fitting). For example, we show person retrieval samples of PCB~\cite{DBLP:conf/eccv/SunZYTW18} in the top-10 ranking list in Fig.~\ref{fig1}. As seen, most of false positive images have similar color information with the query images. Particularly, we remove the color information by making the testing images greyscale. When we evaluate the RGB model (i.e., the model is trained on RGB images) on these greyscale images, mAP decreases by $54.3\%$ ($78.5$ vs. $24.2$), $50.8\%$ ($68.5$ vs. $17.7$) and $44.5\%$ ($61.0$ vs. $16.5$) on Market1501, DukeMTMC-reID and CUHK03, respectively, as shown in Fig.~\ref{fig7} ({\color{orange}orange bars} vs. {\color{new_blue}blue bars}). On one hand, this declares that the color information is important to identify one person. On the other hand, this also implies that the RGB models focus too much on the color information of person images. In this case, using RGB images for person Re-ID may neglect some extra information besides the color information, such as the texture and structure information. 

\begin{figure}
\centering
\includegraphics[width=14cm]{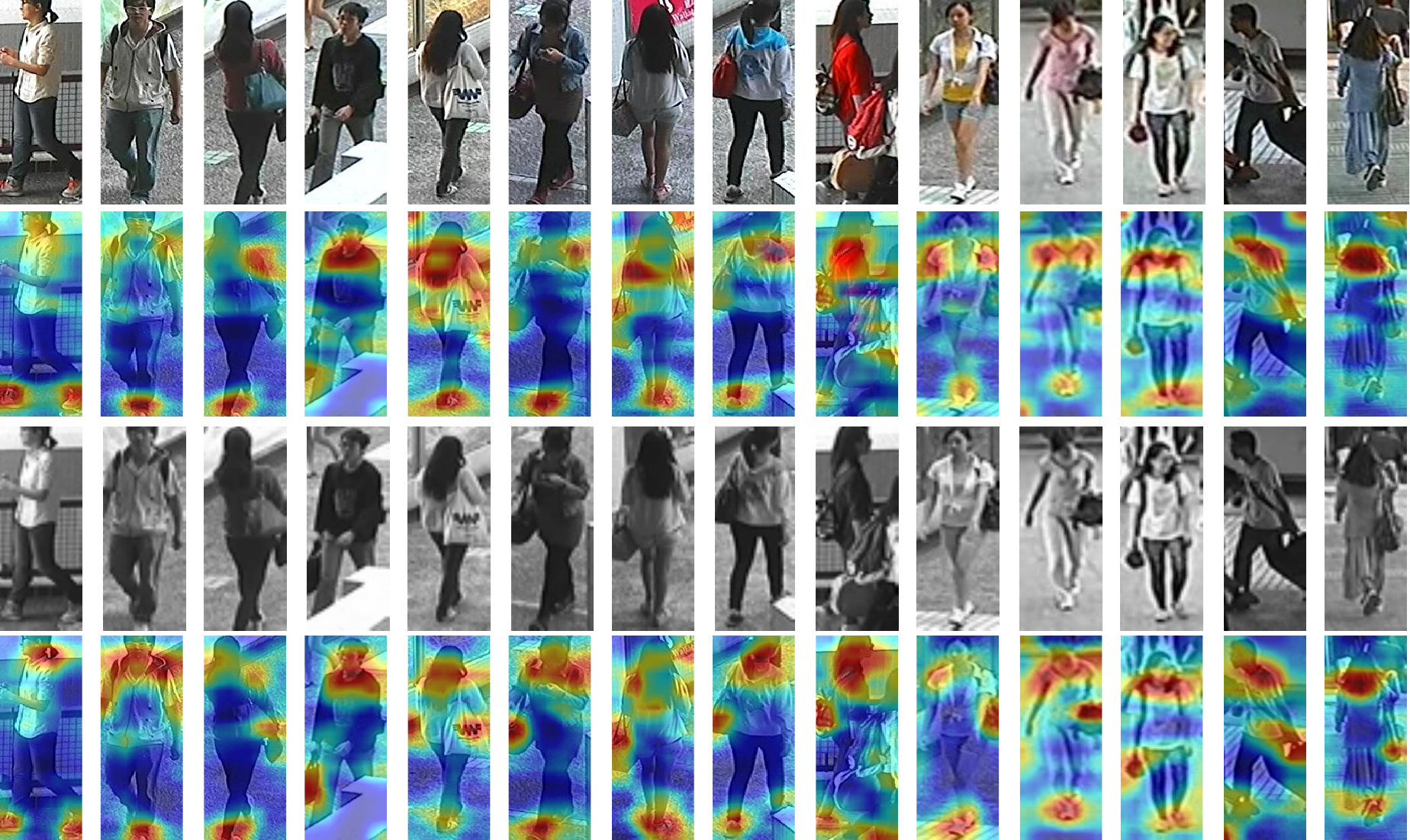}
\caption{Feature response maps of the last convolution layer in the greyscale and RGB models on CUHK03. The greyscale (RGB) model represents that ResNet-50 is trained on greyscale (RGB) images. The response map of each person image is calculated by the mean of all feature maps. First and third rows denote the RGB and greyscale images, respectively, and second and fourth rows are their corresponding feature response maps.   
}
\label{fig8}
\end{figure}
Furthermore, we also observe that RGB and greyscale images for person Re-ID have the complementarity from three different views. First, according to person retrieval samples illustrated in Fig.~\ref{fig2}, by employing greyscale images, some hard query images can achieve better performance than the case of employing RGB images. For example, in Fig.~\ref{fig2}, some query images cannot find any true positive samples in the top-10 ranking list, while using greyscale images can successfully find some. Second, we also train ResNet-50~\cite{DBLP:conf/cvpr/HeZRS16} on greyscale person images (i.e., the greyscale model). The results are shown in Fig.~\ref{fig7} ({\color{grey}grey bars}). For evaluating the greyscale model on the greyscale data, although the performance has a gap with the RGB model tested on RGB images as shown in Fig.~\ref{fig7} ({\color{orange}orange bars}), it has a great improvement compared with the RGB model evaluated on greyscale data ({\color{new_blue}blue bars} in Fig.~\ref{fig7}). For example, the greyscale model significantly gains $33.4\%$ ($79.8$ vs. $46.4$), $35.3\%$ ($66.2$ vs. $30.9$) and $34.1\%$ ($48.9$ vs. $14.8$) in Rank-1 accuracy on Market1501, DukeMTMC-reID and CUHK03, respectively. Thus, we can reasonably deduce that the significant improvements could be from the extra information besides the color information, such as the texture or structure information that may not be sufficiently focused by the RGB model. This further confirms the complementarity of RGB and greyscale models.
Third, according to feature response maps, we observe that greyscale and RGB images can guide the neural network to focus on different regions. Fig.~\ref{fig8} shows feature response maps of greyscale and RGB images in greyscale and RGB models, respectively. As seen, some RGB images neglect the bag regions, while these regions have strong responses in greyscale images.

Considering the complementarity of RGB and greyscale images, our goal is to jointly use greyscale and RGB images to improve the person Re-ID performance by mitigating the color over-fitting issue.
Particularly, unlike the GANs-based methods which generate extra images, the greyscale images in our method can be trivially obtained by converting RGB images as 
\begin{equation}\label{eq04}
\begin{aligned}
Grey(i,j)=0.299\times R(i,j)+0.587\times G(i,j)+0.114\times B(i,j),
  \end{aligned}
\end{equation} where $Grey(i,j)$ denotes the pixel value of greyscale images in the $i$-th row and the $j$-th column. $R$, $G$ and $B$ are three channels in RGB images. As usual, to use ResNet50, we expand greyscale images to three channels with the same pixel values.

\subsection{Two-stream deep framework with RGB-grey information}~\label{RGFFN}
\begin{figure}
\centering
\includegraphics[width=14cm]{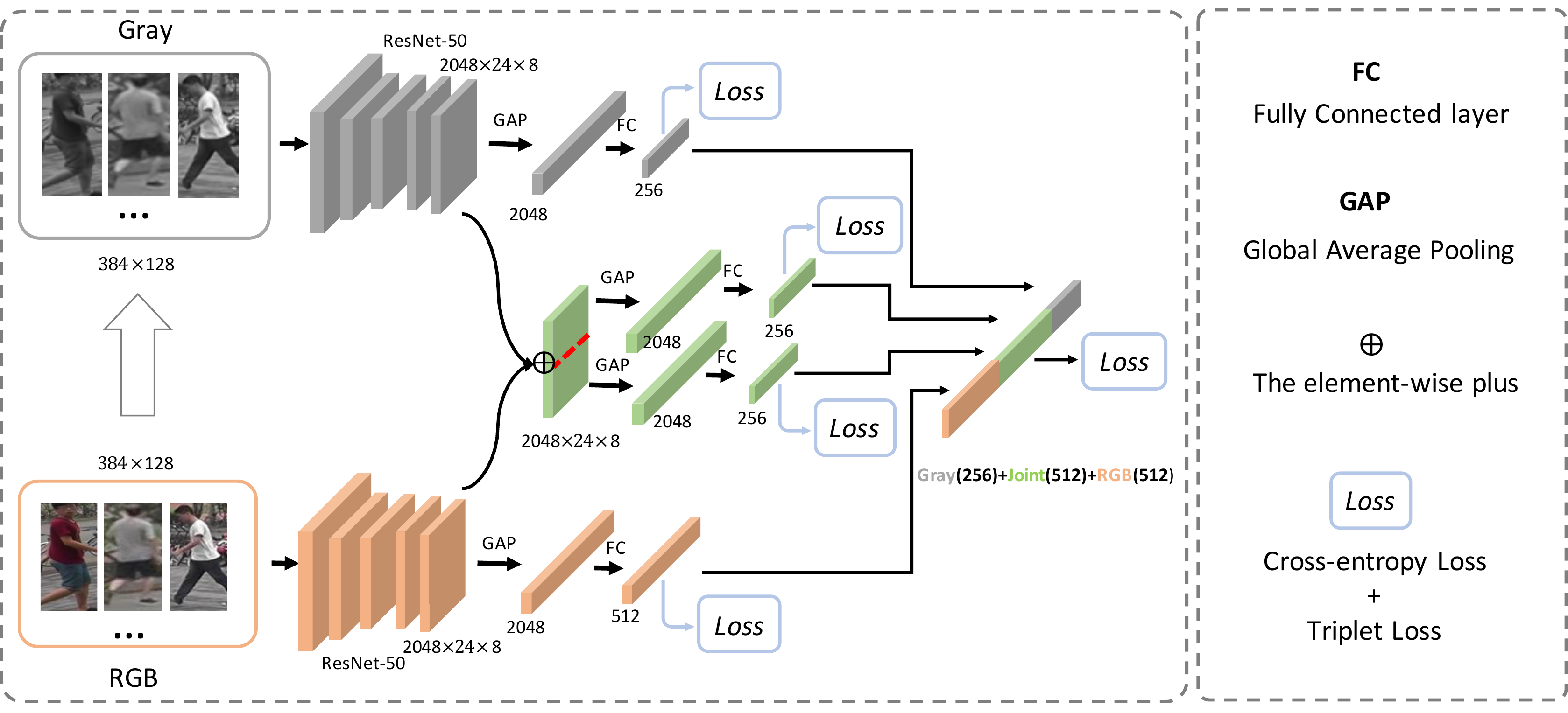}
\caption{Illustration of the proposed two-stream deep framework with RGB-grey information. First, we develop a two-stream network for the RGB and greyscale branches. Then, we fuse the two branches into a joint branch. In particular, in the joint branch, we divide the fusion tensor into two parts. Last, we concatenate the features of all branches as the testing feature for Re-ID. During training, we jointly optimize all branches in the proposed framework by the loss of each branch and the global losses in an end-to-end manner.}
\label{fig3}
\end{figure}

To effectively use greyscale person images, we propose a two-stream deep framework with RGB-grey information for person Re-identification, as illustrated in Fig.~\ref{fig3}. The detailed information is described as follows.

\textbf{The Network Framework.}
In this paper, we develop a two-stream deep learning framework to fuse RGB and greyscale feature representations, whose backbones are the pre-trained ResNet-50 on ImageNet~\cite{DBLP:conf/cvpr/DengDSLL009}. In our framework, we first feed RGB images to the RGB branch, and convert RGB images to greyscale images that are fed into the greyscale branch.  Second, in the last convolutional layer of ResNet-50, we merge the tensors of the greyscale and RGB branches into a new joint branch to connect them. Particularly, for the joint branch, we adopt the part-based scheme to divide the tensor into two parts, whose effectiveness has been validated in the literature~\cite{DBLP:conf/eccv/SunZYTW18,DBLP:conf/mm/WeiZY0T17,DBLP:conf/iccv/SuLZX0T17}. Concretely, given a joint tensor $T_{joint} \in \mathbb{R}^{2048\times 24 \times 8}$, we divide it into two single non-overlapping sub-tensors (i.e., $T_{sub1}$ and $T_{sub2} \in \mathbb{R}^{2048\times 12 \times 8}$). 
Then, the global average pooling (GAP) is employed in all branches. After this, we add a fully connected (FC) layer to form the feature representation of each branch. Finally, we concatenate the features of all branches as the testing feature for Re-ID. In particular, in each branch, we use an independent loss function to train the proposed model simultaneously. Besides, a global loss function is used to fine-tune the concatenated feature, which can further boost the person representation discrimination. 
In this work, we set 256-, 512- and 512-d (dimensional) features for the greyscale, RGB, and joint branches, respectively.


\textbf{Loss Function.}
In person Re-ID, the existing methods construct the classification~\cite{DBLP:conf/cvpr/XiaoLOW16,dai2018batch,zhao2017spindle,li2017person} or discrimination~\cite{cheng2016person,varior2016siamese,yi2014deep,chen2017beyond} tasks to train deep models. In particular, the effectiveness of simultaneously optimizing the classification and discrimination tasks is demonstrated in~\cite{zheng2018discriminatively}. For example, in the discrimination task, the loss function (e.g., contrastive loss~\cite{hadsell2006dimensionality} or triplet loss~\cite{hermans2017defense}) directly calculates the Euclidean distance between two embeddings. In the classification task, the input is independent to each other. But there is implicit relationship between the learned embeddings built by the cross-entropy loss. Thus, we incorporate this joint task to optimize the proposed network. The loss function of the proposed method consists of a cross-entropy loss and a triplet loss with hard sample mining~\cite{hermans2017defense}. For the cross-entropy loss of the classification task, it can be defined as
\begin{equation}\label{eq01}
\begin{aligned}
\mathcal{L}_{\mathrm{Cross}}(X,Y)=-\sum_{i=1}^{N}\sum_{c=1}^{C}\delta(y_{i}-c)\log p(c|x_i),
  \end{aligned}
\end{equation} where $\delta(\cdot)$ is the Dirac delta function. For a sample $x_i$ belonging to the $y_i$-th person class, $p(c|x_i)$ denotes its predicted probability for the $c$-th person class. $N$ is the batch size, and $C$ represents the total number of person classes.

In each batch, we randomly select $P$ persons and each person has $K$ images. $N=P\times K$ is the total number of images in a batch. The triplet loss with hard sample mining can be described as
\begin{equation}\label{eq02}
\begin{aligned}
\mathcal{L}_{\mathrm{Triplet}}(X,Y)=\underbrace{\sum_{i=1}^{P}\sum_{a=1}^{K}}_{one~batch}[m+l(x_{a}^{i})]_{+},
  \end{aligned}
\end{equation} where
\begin{equation}\label{eq03}
\begin{aligned}
l(x_{a}^{i})=\overbrace{\max_{p=1...K}D(f(x_{a}^{i}), f(x_{p}^{i}))}^{hardest~positive}-\overbrace{\min_{\substack{j=1...P\\ n=1,...K \\j\neq i}}D(f(x_{a}^{i}), f(x_{n}^{j}))}^{hardest~negative},
  \end{aligned}
\end{equation} and $m$ denotes the margin. $f(x_{a}^{i})$ is the feature of sample $x_{a}^{i}$ and $D(\cdot,\cdot)$ indicates Euclidean distance. 

In summary, the loss function of each branch can be written as
\begin{equation}\label{eq05}
\begin{aligned}
\mathcal{L}(X,Y)=\mathcal{L}_{\mathrm{Cross}}(X,Y)+\lambda \mathcal{L}_{\mathrm{Triplet}}(X,Y),
  \end{aligned}
\end{equation} where $\lambda$ is the hyper-parameter, which is employed to trade off the two losses. 
In our framework, each branch has an independent loss. Besides, we also utilize a global loss to further fine-tune the concatenated features consisting of the greyscale, RGB and joint features, as shown in Fig.~\ref{fig3}.


\subsection{Discussion}~\label{Discussion}
\textbf{Why does the proposed framework not use the adaptive weight fusion scheme?} 
To obtain better joint tensor, we investigate different tensor fusion schemes in our framework, which include element-wise plus, element-wise multiply, concatenated operation and the adaptive weighting scheme. However, through our experiments, the performance of these tensor fusion schemes does not have enough difference. The main reason is that jointly optimizing the independent loss function of each branch can obtain robust features from all branches. Moreover, the global loss function has been able to automatically adjust the importance of the greyscale, RGB and joint features in the concatenated feature representation. In some sense, this is equivalent to having an adaptive weighting method for the greyscale, RGB and joint branches in the proposed framework.

\textbf{Why does the proposed framework not use the part-based scheme in other branches?} 
In our framework, we only employ the part-based scheme in the joint branch. Since the joint branch is the fused part of the greyscale and RGB branches, using the part-based scheme in the joint branch not only improves the performance of the joint branch, but also helps to improve the generalization ability of the RGB and greyscale branches. This has been validated in our experiments. In practice, we also test the case of using the part-based scheme to the RGB branch. However, it cannot bring a significant improvement with respect to our design.

\begin{figure}
\centering
\subfigure[Our task]{
\includegraphics[width=6cm]{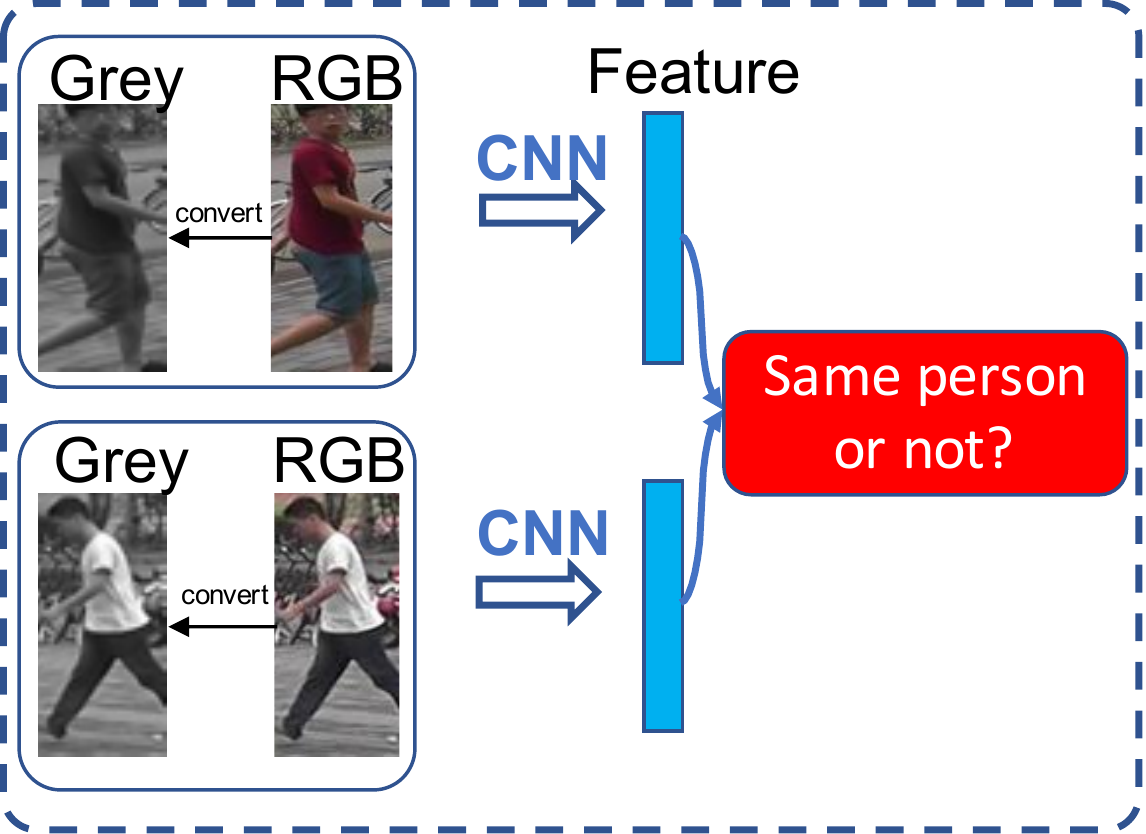}
}
\subfigure[Cross-modality task]{
\includegraphics[width=5.5cm]{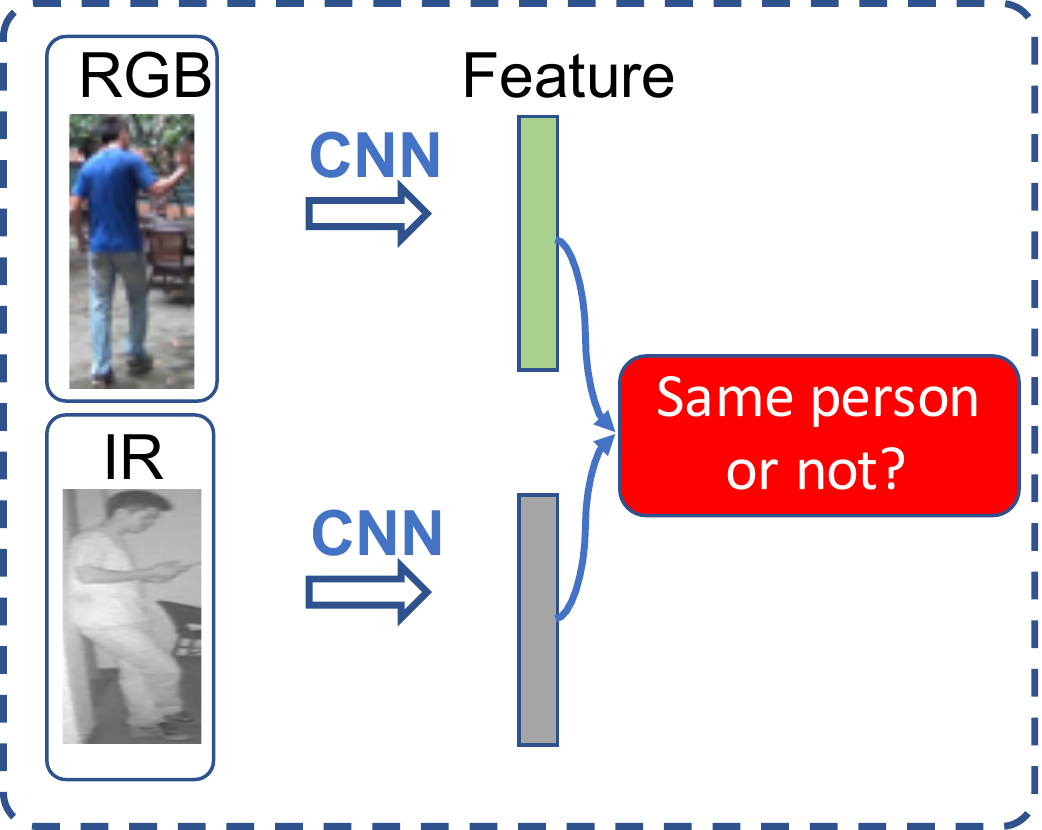}
}
\caption{Comparison between our task and the cross-modality task in~\cite{wu2017rgb,zhang2019dhml,wu2020rgb}. Note that one feature is produced by using ``two'' images (i.e., one RGB image and its corresponding greyscale image) in our task. Differently, one feature is generated by using ``one'' image (i.e., one RGB image or one IR image) in the cross-modality task. Best viewed in color.}
\label{fig11}
\end{figure}

\textbf{The different between the proposed method and the cross-modal method.} 
Our method is fundamentally different from these cross-modality methods in~\cite{wu2017rgb,zhang2019dhml,wu2020rgb}. Concretely, we aim to address the conventional person Re-ID task (i.e., the ``same-modality'' image matching, that is verifying whether two RGB images belong to the same or different persons). Our framework only employs greyscale images to further boost the generalization ability of the extracted features. Differently, the methods in~\cite{wu2017rgb,zhang2019dhml,wu2020rgb} to deal with the ``cross-modality'' task (i.e., verifying whether one RGB image and one IR image belong to the same or different identities), as illustrated in Fig.~\ref{fig11}. Moreover, in our framework, we generate the feature representation for each person by using both its raw RGB image and its corresponding greyscale image. Differently, the cross-modality task extracts IR and RGB image features independently (i.e., one feature is produced from either one IR image or one RGB image), as shown in Fig.~\ref{fig11}. In addition, since our goal is not to match RGB images and greyscale images, these methods in~\cite{wu2017rgb,zhang2019dhml,wu2020rgb} is not suitable to solve our task.

\section{Experiments}\label{s-experiment}
In this part, we first introduce the experimental datasets and settings in Section~\ref{sec:EXP-DS}. Then, we compare the proposed method with baseline and the state-of-the-art methods in Sections~\ref{sec:EXP-CB} and~\ref{sec:EXP-CSM}, respectively. To validate the effectiveness of various components in the proposed framework, we conduct ablation studies in Section~\ref{sec:EXP-AS}. Lastly, we further analyze the
property of the proposed network in Section~\ref{sec:EXP-FA}.
\subsection{Datasets and settings}\label{sec:EXP-DS}
We evaluate our approach on four large-scale datasets: Market1501~\cite{DBLP:conf/iccv/ZhengSTWWT15}, DukeMTMC-reID (Duke)~\cite{ristani2016performance,zheng2017unlabeled}, CUHK03-NP~\cite{DBLP:conf/cvpr/ZhongZCL17} and MSMT17~\cite{wei2018person}. 
 \textbf{Market1501} contains 1,501 persons with 32,668 images from six cameras. Among them, $12,936$ images of 751 identities are used as training set. For evaluation, there are $3,368$ and $19,732$ images in the query set and the gallery set, respectively. \textbf{DukeMTMC-reID} has $1,404$ persons from eight cameras, with $16,522$ training images, $2,228$ queries, and $17,661$ gallery images.
 \textbf{CUHK03-NP} is a new training-testing split protocol for CUHK03. CUHK03~\cite{DBLP:conf/cvpr/LiZXW14} datasets contain two subsets which provide labeled and detected (from a person detector) person images. The detected CUHK03 set includes 7,365 training images, 1,400 query images and 5,332 gallery images. The labeled set contains 7,368 training, 1,400 query and 5,328 gallery images respectively. The new protocol in~\cite{DBLP:conf/cvpr/ZhongZCL17} splits the training and testing sets into 767 and 700 identities.
 \textbf{MSMT17} is collected from a 15-camera network deployed on campus. The training set contains $32,621$ images of $1,041$ identities. For evaluation, $11,659$ and $82,161$ images are used as query and gallery images, respectively. For all datasets, we employ CMC accuracy and mAP for Re-ID evaluation~\cite{DBLP:conf/iccv/ZhengSTWWT15}. On Market1501, there are single- and multi-query evaluation protocols. We use the more challenging single-query protocol in our experiments.

 In the training process, we set $P$ and $K$ to $32$ and $4$, respectively. The margin of triplet loss, $m$, is $0.3$. The proposed model is trained with the SGD optimizer in a total of $300$ epochs. The initial learning rate is set to $0.01$, and decreases to $0.001$ and $0.0001$ at the $100$-th and $200$-th epochs, respectively. During training, the input images are resized to $384 \times 128$ and then pre-processed by random horizontal flip, normalization, and random erasing~\cite{zhong2017random}. Particularly, all experiments in this paper utilize the same setting on all datasets.

Note that the  \textbf{baseline} model in this experiment represents the pre-trained ResNet-50~\cite{DBLP:conf/cvpr/HeZRS16} on ImageNet~\cite{DBLP:conf/cvpr/DengDSLL009} with the loss in Eq.~(\ref{eq05}). Particularly, for fair comparison, we set different feature dimensions in the baseline model for different experiments. 
Baseline-grey and Baseline-rgb denote that the baseline model is trained on greyscale and RGB images, respectively. 

\subsection{Comparison with the baseline}\label{sec:EXP-CB}
\begin{figure}
\centering
\includegraphics[width=14cm]{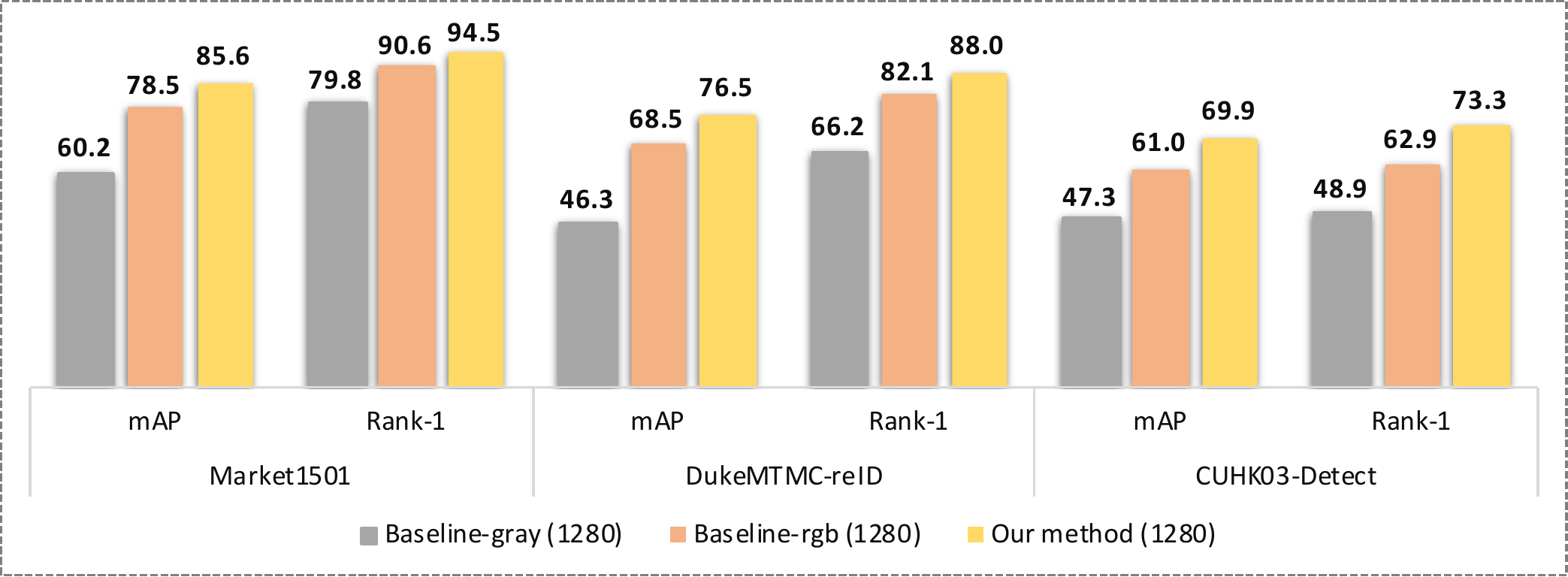}
\caption{Comparison of the baseline and our method on Market1501, DukeMTMC-reID and CUHK03-NP, respectively. Particularly, the baseline model is trained on grey and RGB images, respectively. Note that all methods extract 1280-dimensional features.}
\label{fig4}
\end{figure}

In this section, we compare the proposed methods and the baseline method, as shown in Fig.~\ref{fig4}. 1280-d features are extracted for all methods. Particularly, we conduct the baseline method on greyscale and RGB images, respectively. First, fusing RGB and greyscale features can achieve better results than the features of RGB or greyscale images alone. As seen in Fig.~\ref{fig4}, when compared with Baseline-rgb, our method can significantly improve mAP by $7.1\%$ ($85.6$ vs. $78.5$), $8.0\%$ ($76.5$ vs. $68.5$) and $8.9\%$ ($69.9$ vs. $61.0$) on Market1501, DukeMTMC-reID and CUHK03-NP-detect, respectively. 
This validates the effectiveness of the proposed two-stream deep framework. Second, using greyscale images only to conduct the Re-ID task has poor performance compared with employing RGB images. This can be excerpted because the color information is also important to identify a person. Therefore, the proposed framework also utilizes the RGB branch to extract color information.

\subsection{Comparison with the-state-of-the-art methods}\label{sec:EXP-CSM}
We compare the proposed methods with the state-of-art methods including attribute-based person Re-ID (A$^3$M~\cite{DBLP:conf/mm/HanGZZ18},AANet~\cite{tay2019aanet}, CA$^3$NET~\cite{DBLP:conf/mm/LiuZXXZ18}, MHN-6~\cite{chen2019mixed}, CAMA~\cite{yang2019towards}, IANet~\cite{hou2019interaction}), segmentation-based methods (MGCAM~\cite{DBLP:conf/cvpr/Song0O018}, MaskReID~\cite{qi2018maskreid}, SPReID~\cite{DBLP:conf/cvpr/KalayehBGKS18}), GANs-based methods (Pose-transfer~\cite{DBLP:conf/cvpr/LiuNYZCH18}, CamStyle+RE~\cite{DBLP:conf/cvpr/Zhong0ZL018}, JDGL~\cite{zheng2019joint}), part-based methods (PDC~\cite{DBLP:conf/mm/WeiZY0T17}, GLAD~\cite{DBLP:conf/iccv/SuLZX0T17}, PCB+RPP~\cite{DBLP:conf/eccv/SunZYTW18}, P$^2$-Net~\cite{guo2019beyond}), and attention-based methods (Mancs~\cite{DBLP:conf/eccv/WangZHLW18}, CASN~\cite{zheng2018re}) on Market1501, Duke, CUHK03-NP and MSMT17, respectively. The experimental results are reported in Tables~\ref{tab-1},~\ref{tab-2} and~\ref{tab-3}. Compared with A$^3$M~\cite{DBLP:conf/mm/HanGZZ18}, CA$^3$NET~\cite{DBLP:conf/mm/LiuZXXZ18}, MGCAM~\cite{DBLP:conf/cvpr/Song0O018}, and SPReID~\cite{DBLP:conf/cvpr/KalayehBGKS18}, which use extra information such as person attribute or the image-segmentation information, the proposed method achieves considerable improvement on all datasets. In particular, compared with SPReID, which uses the ResNet-152 as the backbone, our method still can improve mAP by $2.2\%$ ($85.6$ vs. $83.4$) and $3.2\%$ ($76.5$ vs. $73.3$) on Market-1501 and Duke, respectively. Although Pose-transfer~\cite{DBLP:conf/cvpr/LiuNYZCH18} and CamStyle+RE~\cite{DBLP:conf/cvpr/Zhong0ZL018} generate many extra images to alleviate the model over-fitting, they still show an inferior performance to our method. For example, on CUHK03-Detect and CUHK03-Label, the gap between Pose-transfer and our method is about $30\%$ in both mAP and  Rank-1 accuracy. The effectiveness of part-based methods (PDC~\cite{DBLP:conf/mm/WeiZY0T17}, GLAD~\cite{DBLP:conf/iccv/SuLZX0T17}, PCB+RPP~\cite{DBLP:conf/eccv/SunZYTW18}) and attention-based methods (Mancs~\cite{DBLP:conf/eccv/WangZHLW18}, CASN~\cite{zheng2018re}) has been validated in recent years. Compared with these methods, our method can still achieve the competitive results on all datasets. Particularly, on MSMT17, the proposed method improves PCB by $11.8\%$ ($55.0$ vs. $43.2$) and $8.3\%$ ($78.6$ vs. $70.3$) in mAP and Rank-1 accuracy, respectively. These results well demonstrate the effectiveness and advantage of the proposed method by smartly integrating the greyscale and color information of person images. 

\renewcommand{\cmidrulesep}{0mm} 
\setlength{\aboverulesep}{0mm} 
\setlength{\belowrulesep}{0mm} 
\setlength{\abovetopsep}{0cm}  
\setlength{\belowbottomsep}{0cm}
\setlength{\tabcolsep}{12pt}
\begin{table}[htbp]
  \centering
  \caption{Comparison with the state-of-the-art methods on Market1501 and DukeMTMC-reID, respectively. ``-'' denotes that the result is not provided. The best performance is shown in \textbf{bold}.}
    \begin{tabular}{|c|cc|cc|}
    \toprule
    \midrule
    \multirow{2}[1]{*}{Mehods} & \multicolumn{2}{c|}{Market1501} & \multicolumn{2}{c|}{DukeMTMC-reID} \\
\cmidrule{2-5}          & mAP   & Rank-1 & mAP   & Rank-1 \\
    \midrule
    A$^3$M~\cite{DBLP:conf/mm/HanGZZ18}   & 69.0  & 86.5  & 70.2  & 84.6 \\
    CA$^3$NET~\cite{DBLP:conf/mm/LiuZXXZ18} & 80.0  & 93.2  & 70.2  & 84.6 \\
    AANet~\cite{tay2019aanet} & 82.5  & 93.9  & 72.6  & 86.4 \\
    MHN-6 (IDE)~\cite{chen2019mixed} & 83.6 & 93.6 &  75.2 & 87.5\\
    CAMA~\cite{yang2019towards} & 84.5 &  \textbf{94.7} & 72.9 &  85.8 \\
    IANet~\cite{hou2019interaction} & 83.1 & 94.4 & 73.4 & 87.1 \\
    \midrule
    MGCAM~\cite{DBLP:conf/cvpr/Song0O018} & 74.3  & 83.8  &   -    & - \\
    MaskReID~\cite{qi2018maskreid} & 75.4  & 90.4  &   61.9    & 78.9 \\
    SPReID~\cite{DBLP:conf/cvpr/KalayehBGKS18} & 83.4  & 93.7  & 73.3  & 86.0 \\
    P$^2$-Net~\cite{guo2019beyond} & 83.4 & 94.0 & 70.8 & 84.9\\
    \midrule
    Pose-transfer~\cite{DBLP:conf/cvpr/LiuNYZCH18} & 68.9  & 87.7  & 56.9  & 78.5 \\
    CamStyle+RE~\cite{DBLP:conf/cvpr/Zhong0ZL018} & 71.6  & 89.5  & 57.6  & 78.3 \\
    \midrule
    PDC~\cite{DBLP:conf/mm/WeiZY0T17}   & 63.4  & 84.1  &    -   &  -\\
    GLAD~\cite{DBLP:conf/iccv/SuLZX0T17}  & 73.9  & 89.9  &    -   &  -\\
    PCB~\cite{DBLP:conf/eccv/SunZYTW18}
   & 77.3  & 92.4  & 65.3  & 81.9 \\
    PCB+RPP~\cite{DBLP:conf/eccv/SunZYTW18}
 & 81.6  & 93.8  & 69.2  & 83.3 \\
    \midrule
    Mancs~\cite{DBLP:conf/eccv/WangZHLW18} & 82.3  & 93.1  & 71.8  & 84.9 \\
    CASN~\cite{zheng2018re}  & 82.8  & 94.4  & 73.7  & 87.7 \\
    \midrule
    ~~GreyReID (ours)~~ & \textbf{85.6} & 94.5 & \textbf{76.5} & \textbf{88.0} \\
    \bottomrule
    \end{tabular}%
  \label{tab-1}%
\end{table}%

\begin{table}[htbp]
  \centering
  \caption{Comparison with the state-of-the-art methods on CUHK03-NP. The results are reported on both the labeled and detected CUHK03 set. ``-'' denotes that the result is not provided. The best performance is shown in \textbf{bold}.}
    \begin{tabular}{|c|cc|cc|}
    \toprule
    \midrule
    \multirow{2}[1]{*}{Methods} & \multicolumn{2}{c|}{CUHK03-Detect} & \multicolumn{2}{c|}{CUHK03-Label} \\
\cmidrule{2-5}          & mAP & Rank-1   & mAP & Rank-1 \\
    \midrule
    MGCAM~\cite{DBLP:conf/cvpr/Song0O018} & 46.9  & 46.7  & 50.2  & 50.1 \\
    Pose-transfer~\cite{DBLP:conf/cvpr/LiuNYZCH18} & 38.7  & 41.6  & 42.0  & 45.1 \\
    P$^2$-Net~\cite{guo2019beyond} & 64.2 & 71.6 &  69.2 & 75.8\\
    PCB~\cite{DBLP:conf/eccv/SunZYTW18}   & 54.2  & 61.3  &  -     & - \\
    PCB+RPP~\cite{DBLP:conf/eccv/SunZYTW18} & 56.7  & 62.8  &  -     & - \\
    CAMA~\cite{yang2019towards} & 64.2 &  66.6  & 66.5 &  70.1 \\
     MHN-6 (IDE)~\cite{chen2019mixed} & 61.2 & 67.0 & 65.1 & 69.7 \\
    Mancs~\cite{DBLP:conf/eccv/WangZHLW18} & 60.5  & 65.5  & 63.9  & 69.0 \\
    CASN~\cite{zheng2018re}  & 64.4  & 71.5  & 68.0  & 73.7 \\
    \midrule
    GreyReID (ours) & \textbf{69.9} & \textbf{73.3} & \textbf{73.9} & \textbf{76.6} \\
    \bottomrule
    \end{tabular}%
  \label{tab-2}%
\end{table}%

\begin{table}[htbp]
  \centering
  \caption{Comparison with the state-of-the-art methods on MSMT17. In this table, we report Rank-1, 5, 10 of CMC accuracy and mAP. ``-'' denotes that the result is not provided. The best performance is shown in \textbf{bold}.}
    \begin{tabular}{|c|cccc|}
    \toprule
    \midrule
    \multirow{2}[1]{*}{Methods} & \multicolumn{4}{c|}{MSMT2017} \\
\cmidrule{2-5}          & mAP   & Rank-1 & Rank-5 & Rank-10 \\
    \midrule
    PDC~\cite{DBLP:conf/mm/WeiZY0T17}   & 29.7  & 58.0  & 73.6  & 79.4 \\
    GLAD~\cite{DBLP:conf/iccv/SuLZX0T17}  & 34.0  & 61.4  & 76.8  & 81.6 \\
    PCB~\cite{DBLP:conf/eccv/SunZYTW18}   & 43.2  & 70.3  & 82.9  & 86.7 \\
     IANet~\cite{hou2019interaction} & 46.8 & 75.5 &  85.5 & 88.7 \\
    JDGL~\cite{zheng2019joint} & 52.3 & 77.2 & 87.4 & 90.5\\
\cmidrule{1-5}    GreyReID (ours) & \textbf{55.0}  & \textbf{78.6}  & \textbf{88.3}  & \textbf{91.2} \\
    \bottomrule
    \end{tabular}%
  \label{tab-3}%
\end{table}%

\subsection{Ablation studies}\label{sec:EXP-AS}
\setlength{\tabcolsep}{6pt}

\begin{table*}[htbp]
  \centering
  \caption{Performance of the feature combination of different branches on Market1501, DukeMTMC-reID (Duke), CUHK03-NP-Detect (CHUK03) and MSMT17, respectively. Note that in this table, ``+'' denotes the concatenated operation. Two-part (one-part) indicates the proposed framework with (without) the part-based scheme in the joint branch. The best performance is \textbf{bold}.}
    \begin{tabular}{|c|c|C{0.3cm}C{1.1cm}|C{0.3cm}C{1.1cm}|C{0.3cm}C{1.1cm}|C{0.3cm}C{1.1cm}|}
    \toprule
    \midrule
    \multirow{2}[2]{*}{} & \multirow{2}[1]{*}{Different branches} & \multicolumn{2}{c|}{Market1501} & \multicolumn{2}{c|}{Duke} & \multicolumn{2}{c|}{CUHK03} & \multicolumn{2}{c|}{MSMT17} \\
\cmidrule{3-10}          &       & mAP   & Rank-1 & mAP   & Rank-1 & mAP   & Rank-1 & mAP   & Rank-1 \\
    \midrule
    \multirow{3}[1]{*}{Baseline} & Baseline-grey(256) & 57.0  & 78.7  & 43.8  & 63.7  & 41.2  & 42.5  & 26.1  & 52.1 \\
          & Baseline-rgb(512) & 77.8  & 90.6  & 68.4  & 83.0  & 58.0  & 61.4  & 44.3  & 69.6 \\
          & Baseline-grey+rgb & \textbf{81.8}  & \textbf{91.9}  & \textbf{70.7}  & \textbf{84.3}  & \textbf{62.8}  & \textbf{65.5}  & \textbf{49.0}  & \textbf{74.1} \\
    \midrule
    \midrule
    \multirow{8}[1]{*}{One-part} & Grey(256) & 59.0  & 80.1  & 46.2  & 66.4  & 44.7  & 46.8  & 26.5  & 52.8 \\
          & RGB(512) & 79.2  & 90.7  & 70.0  & 84.1  & 61.0  & 64.6  & 45.8  & 70.4 \\
          & Joint(512) & 82.4  & 92.4  & 71.8  & 85.1  & 64.1  & 67.6  & 50.6  & 75.0 \\
          & Grey+Joint & 80.6  & 92.1  & 68.5  & 82.2  & 62.4  & 66.4  & 47.3  & 73.5 \\
          & RGB+Joint & 82.5  & 92.3  & 72.8  & \textbf{85.4}  & 64.7  & 68.4  & 50.7  & 74.5 \\
          & Grey+RGB & 83.1  & 92.8  & 72.6  & 85.0  & 65.6  & 69.1  & 50.7  & 75.5 \\
          & Grey+RGB+Joint & \textbf{83.5}  & \textbf{93.0}  & \textbf{73.1}  & 85.2  & \textbf{66.2}  & \textbf{69.6}  & \textbf{51.6}  & \textbf{75.9} \\
    \midrule
    \midrule
    \multirow{8}[1]{*}{Two-part} & Grey(256) & 59.5  & 79.8  & 46.2  & 67.9  & 46.2  & 48.6  & 27.2  & 53.1 \\
          & RGB(512) & 80.0  & 91.3  & 71.0  & 84.4  & 62.8  & 67.2  & 46.2  & 70.8 \\
          & Joint(512) & 84.3  & \textbf{94.2}  & 74.9  & 87.0  & 67.9  & 71.4  & 53.2  & 77.4 \\
          & Grey+Joint & 83.5  & 93.6  & 73.2  & 86.7  & 67.3  & 71.1  & 51.9  & 76.9 \\
          & RGB+Joint & 85.0  & 94.1  & 75.8  & \textbf{87.8}  & 69.1  & 72.4  & 53.8  & 77.3 \\
          & Grey+RGB & 83.7  & 93.4  & 73.9  & 87.3  & 67.7  & 71.4  & 52.0  & 76.6 \\
          & Grey+RGB+Joint & \textbf{85.3}  & 94.1  & \textbf{75.9}  & \textbf{87.8}  & \textbf{69.6}  & \textbf{73.1}  & \textbf{54.5}  & \textbf{78.4} \\
    \bottomrule
    \end{tabular}%
  \label{tab-4}%
\end{table*}%

To adequately validate the effectiveness of the proposed method, we remove the global loss to train the model with the one-part scheme (i.e., without the part-based scheme in the proposed network) and the two-part scheme, respectively. We report the feature combination results of different branches on Market1501, Duke, CUHK03 and MSMT17, as shown in Table~\ref{tab-4}. ``Baseline-grey'' and ``Baseline-rgb'' denote that the baseline model is trained on greyscale and RGB images, respectively. For a fair comparison, we set the baseline model to have the same feature dimension with the RGB and greyscale branches in our framework, respectively. Moreover, we concatenate the features of ``Baseline-grey'' and ``Baseline-rgb'' to generate new features (Baseline-grey+rgb'') to conduct person Re-ID. Besides, we also validate the efficacy of the loss function in our framework, as reported in Tables~\ref{tab-5} and \ref{tab-7}.

\textbf{Effectiveness of each branch in the proposed network.} 
In Table~\ref{tab-4}, ``Baseline-grey+rgb'' improves ``Baseline-rgb'' by $4.0\%$ ($81.8$ vs. $77.8$), $2.3\%$ ($70.7$ vs. $68.4$), $4.8\%$ ($62.8$ vs. $58.0$) and $4.7\%$ ($49.0$ vs. $44.3$) in mAP on Market1501, Duke, CUHK03 and MSMT17, respectively. Besides, in our framework, the concatenated feature of the RGB and greyscale branches (``Grey+RGB'') can also enhance the performance when compared with the single RGB branch (``RGB'') for both the one-part and two-part schemes. This demonstrates the effectiveness of greyscale branch in the proposed framework. 
Moreover, the joint branch can achieve better performance than both the greyscale and RGB branches. This further confirms the effectiveness of fusing the RGB and greyscale branches. Particularly, compared with ``Joint'', ``Grey+Joint'' is slightly poorer in all experiments. This is because i) the joint feature has contained the greyscale information; ii) although the greyscale information is important for Re-ID, it also exists some noises.

\textbf{Effectiveness of the part-based scheme.}
As shown in Table~\ref{tab-4}, using the part-based scheme not only improves the performance of the joint branch, but also enhances the generalization ability of the RGB and greyscale branches on most datasets. For example, for Gery+RGB+Joint, the two-part method can gain $1.8\%$ ($85.3$ vs. $83.5$), $2.8\%$  ($75.9$ vs. $73.1$), $3.4\%$ ($69.6$ vs. $66.2$) and $2.9\%$ ($54.5$ vs. $51.6$) in mAP on Market1501, Duke, CUHK03 and MSMT17, respectively.

\textbf{Effectiveness of joint learning framework.}
First, jointly learning RGB and greyscale task is effective to improve the performance of both the greyscale and RGB branches. For example, both ``Grey'' and ``RGB'' have some improvements when compared with ``Baseline-grey'' and ``Baseline-rgb'', respectively. For the one-part scheme, compared with ``Baseline-rgb'', the performance of the RGB branch in our framework can gain $1.4\%$ ($79.2$ vs. $77.8$), $1.6\%$ ($70.0$ vs. $68.4$), $3.0\%$ ($61.0$ vs. $58.0$) and $1.5\%$ ($45.8$ vs. $44.3$) in mAP on Market1501, Duke, CUHK03 and MSMT17, respectively. Moreover, when using the two-part scheme in our framework, the improvements become more significant. Second, through concatenating the features of all branches, we can obtain the more robust feature representation for Re-ID. As shown in Table~\ref{tab-4}, ``Grey+RGB+Joint'' consistently outperforms other feature combinations on most datasets.

\textbf{Evaluation of the loss function.}
To validate the effectiveness of jointly optimizing the classification and discrimination tasks, we separately use the cross-entropy loss and the triplet loss to train the proposed network on CUHK03, as reported in Table~\ref{tab-5}. As seen, the joint loss functions can achieve the best performance. Besides, we also validate the effectiveness of using the independent loss in all branches. We remove the losses of all branches and the global loss from the proposed framework respectively, which are ``w/o branch loss'' and ``w/o global loss'' in Table~\ref{tab-7}. As seen, ``w/o branch loss'' has the poorest performance on all datasets. This confirms that the loss in each branch is crucial to fully optimize the proposed network. In addition, using global loss can further boost the robustness of the concatenated feature.


\setlength{\tabcolsep}{12pt}
\begin{table}[htbp]
  \centering
  \caption{Evaluation of different components in the loss function on  Market1501 and DukeMTMC-reID.}
    \begin{tabular}{|c|cc|cc|}
    \toprule
    \midrule
    \multirow{2}[1]{*}{Loss Function} & \multicolumn{2}{c|}{Market1501} & \multicolumn{2}{c|}{DukeMTMC-reID} \\
\cmidrule{2-5}          & mAP   & Rank-1 & mAP   & Rank-1 \\
    \midrule
    only $l_{triplet}$ in Eq.~(\ref{eq05}) & 75.6  & 89.6  & 66.1  & 81.7 \\
    \midrule
    only $l_{cross}$ in Eq.~(\ref{eq05}) & 83.8  & 93.7  & 73.4  & 86.7 \\
    \midrule
         $l_{cross}+l_{triplet}$ & \textbf{85.6} & \textbf{94.5} & \textbf{76.5} & \textbf{88.0} \\
        \bottomrule
    \end{tabular}%
  \label{tab-5}%
 \vspace*{-10pt}%
\end{table}%

\begin{table}[htbp]
  \centering
  \caption{Evaluation of the branch and global losses in our framework on Market1501, DukeMTMC-reID (Duke), CHUK03-NP-Detect and MSMT17, respectively.  ``w/o global loss''  (``w/o branch loss'') denotes that our framework without the global loss (the losses of all branches). The best performance is \textbf{bold}.}
    \begin{tabular}{|c|cc|cc|}
    \toprule
    \midrule
    \multirow{2}[1]{*}{Different  losses} & \multicolumn{2}{c|}{Market1501} & \multicolumn{2}{c|}{Duke} \\
\cmidrule{2-5}          & mAP   & Rank-1 & mAP   & Rank-1 \\
    \midrule
    w/o branch loss & 79.1  & 91.1  & 69.5  & 82.9 \\
    w/o global loss & 85.3  & 94.1  & 75.9  & 87.8 \\
    \midrule
    GreyReID (ours) & \textbf{85.6} & \textbf{94.5} & \textbf{76.5} & \textbf{88.0} \\
    \midrule
    \midrule
    \multirow{2}[1]{*}{Different  losses} & \multicolumn{2}{c|}{CUHK03-NP-Detect} & \multicolumn{2}{c|}{MSMT17} \\
\cmidrule{2-5}          & mAP   & Rank-1 & mAP   & Rank-1 \\
    \midrule
    w/o branch loss & 58.4  & 60.9  & 44.8  & 69.8 \\
    w/o global loss & 69.6  & 73.1  & 54.5  & 78.4 \\
    \midrule
    GreyReID (ours) & \textbf{69.9} & \textbf{73.3} & \textbf{55.0} & \textbf{78.6} \\
    \bottomrule
    \end{tabular}%
  \label{tab-7}%
\end{table}%

\subsection{Further analysis}\label{sec:EXP-FA}

\begin{figure}
\centering
\includegraphics[width=10cm, height=6cm]{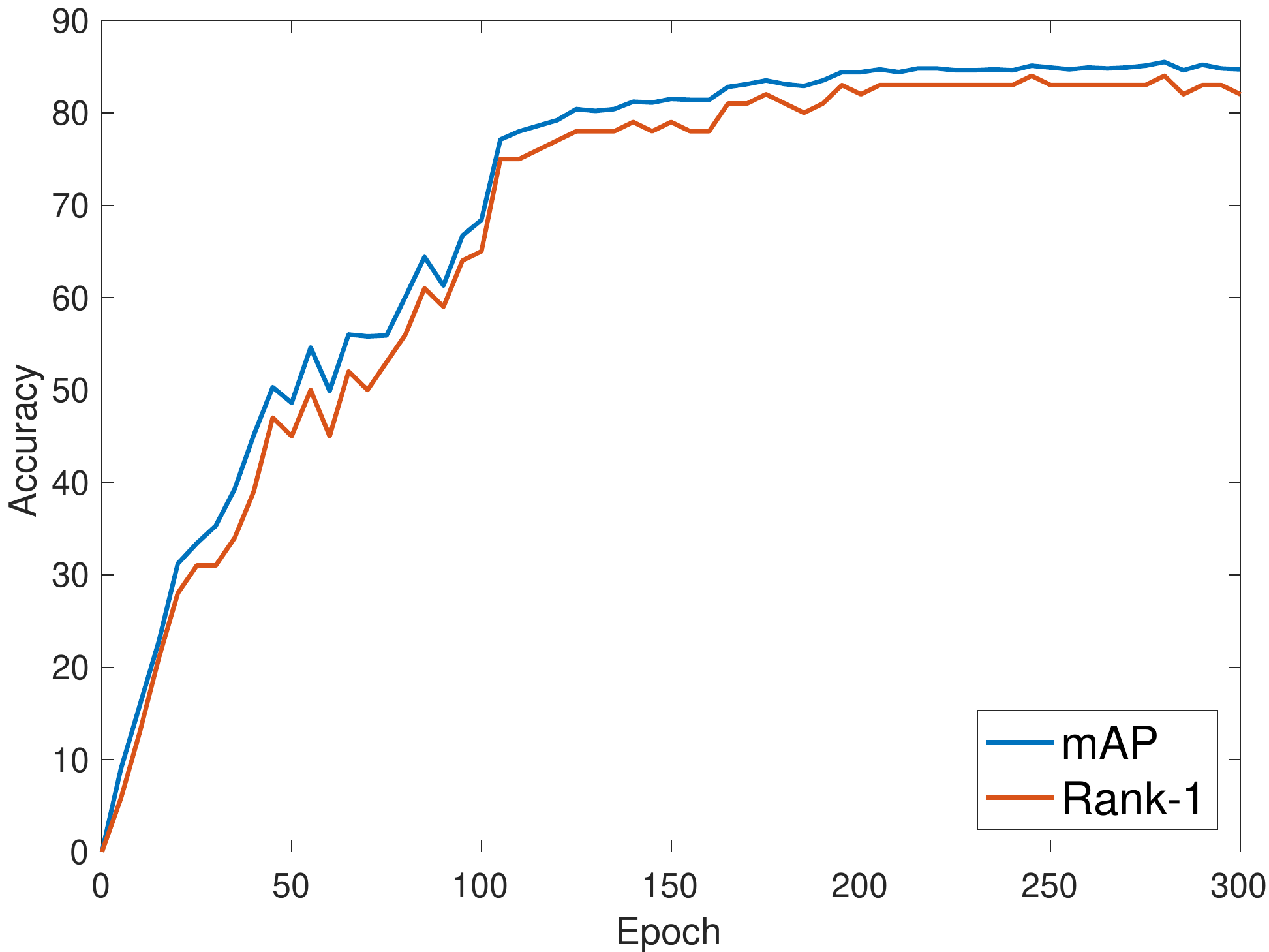}
\caption{Convergence curves of the proposed method on MSMT17.}
\label{fig10}
\end{figure}

\textbf{Algorithm Convergence.}
To investigate the convergence of our algorithm, we record the mAP and Rank-1 accuracy of the proposed method during the iteration on a validated set of MSMT17in Fig.~\ref{fig10}. As seen, we can observe that our proposed method can almost converge after 200 epochs.

\textbf{The sensitiveness of parameter.}
We conduct the experiment with various $\lambda$ in Eq.~(\ref{eq05}) to validate the sensitiveness of the parameter, as reported in Table~\ref{tab06}. According to the experimental results, we observe that i) using cross-entropy loss only can obtain better performance than using triplet loss only. This demonstrates the importance of cross-entropy loss in our case. ii) Using them together can consistently bring more improvement by increasing the value of $\lambda$. However, when giving too much attention on the triplet loss (i.e., $\lambda > 1$), the performance will decrease.
\begin{table}[htbp]
  \centering
  \caption{Evaluation on various $\lambda$ on Market1501 and DukeMTMC-reID.}
    \begin{tabular}{|c|cc|cc|}
    \toprule
    \midrule
    \multirow{2}[1]{*}{$\lambda$} & \multicolumn{2}{c|}{Market1501} & \multicolumn{2}{c|}{DukeMTMC-reID} \\
\cmidrule{2-5}          & mAP   & Rank-1 & mAP   & Rank-1 \\
    \midrule
    only $l_{triplet}$ & 75.6  & 89.6  & 66.1  & 81.7 \\
    \midrule
    only $l_{cross}$ ($\lambda=0$) & 83.8  & 93.7  & 73.4  & 86.7 \\
    $\lambda=$0.2   & 83.9  & 93.4  & 74.2  & 86.2 \\
    $\lambda=$0.5   & 84.7  & 93.9  & 75.2  & 87.4 \\
    $\lambda=$0.8   & 85.4  & 94.3  & 75.9  & 87.0 \\
    $\lambda=$1.0     & \textbf{85.6} & \textbf{94.5} & \textbf{76.5} & \textbf{88.0} \\
    $\lambda=$1.5   & 85.5  & 94.1  & 75.7  & 86.9 \\
    $\lambda=$2.0     & 84.6  & 93.3  & 74.4  & 87.1 \\
    \bottomrule
    \end{tabular}%
  \label{tab06}%
\end{table}%

\begin{figure}
\centering
\subfigure[$\#$image]{
\includegraphics[width=6cm]{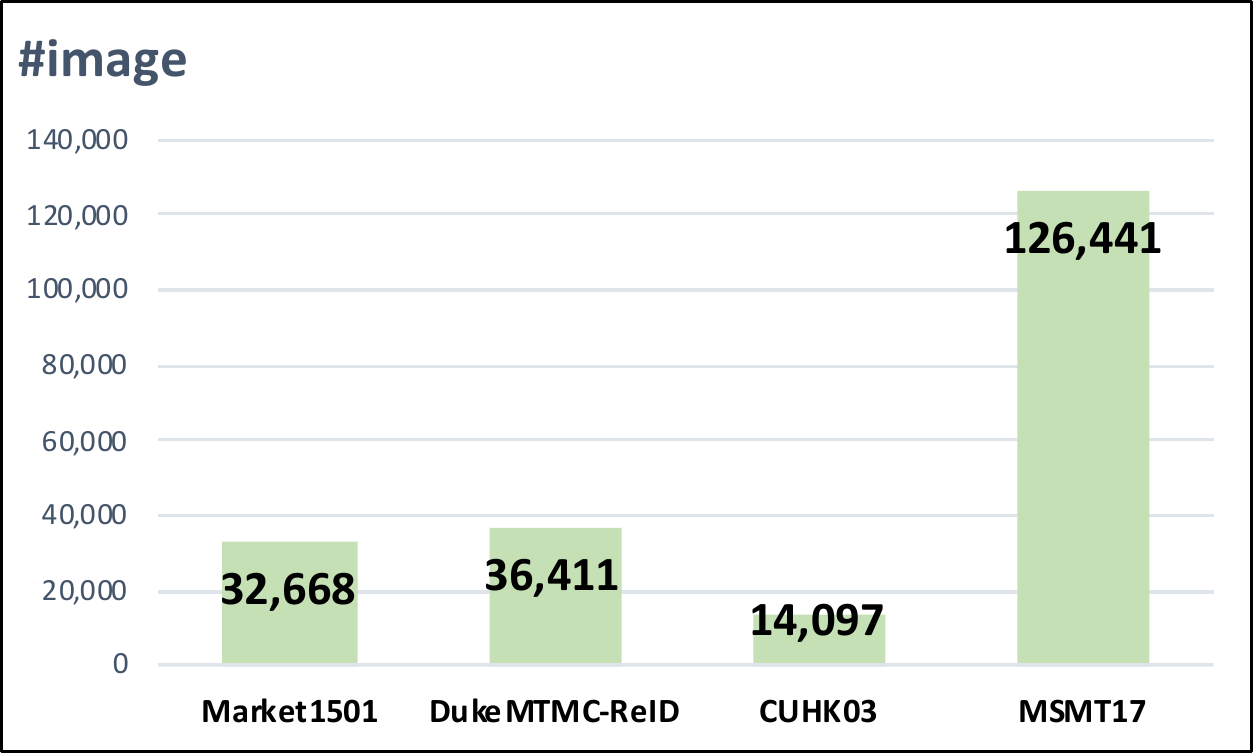}
}
\subfigure[$\#$camera]{
\includegraphics[width=6cm]{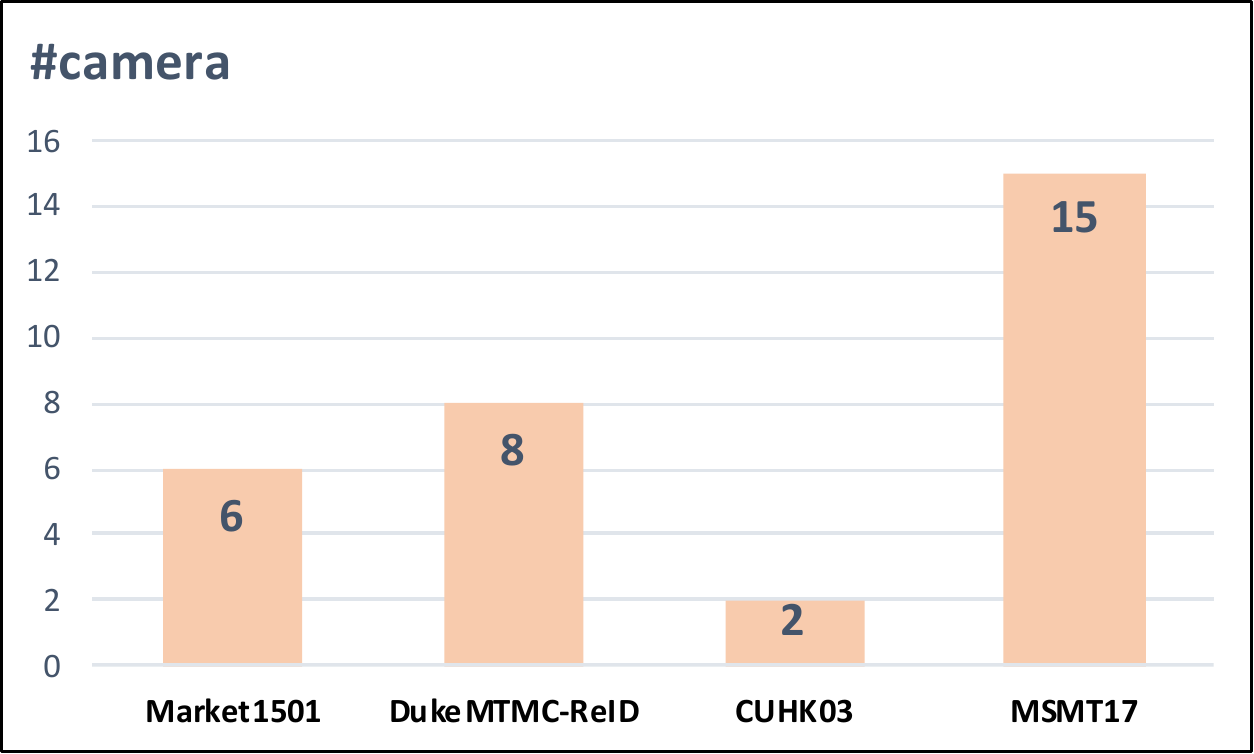}
}
\caption{The numbers of cameras and images on the four benchmark datasets.}
\label{fig12}
\end{figure}

\textbf{The evaluation of different inputs in the proposed framework.}
To further validate the efficacy of greyscale images, we utilize the RGB-RGB pair and the RGB-grey pair as the input of the proposed network, respectively. Note that all settings are the same in this experiment. The experimental results are reported in Table~\ref{tab-8}. As seen, using the RGB-grey pair as input can achieve a consistent improvement on all datasets. For example, on existing largest person Re-ID dataset (i.e., MSMT17), the RGB-grey pair gains $1.4\%$ ($55.0$ vs. $53.6$) and $1.3\%$ ($78.6$ vs. $77.3$) in mAP and Rank-1 accuracy.
Please note that our work consists of two key contributions:
First, compared with the baseline in Table~\ref{tab-8}, which only has a RGB-image input (i.e., without using either the proposed modules of our framework or the greyscale-image input), our framework, regardless of using a RGB-Grey pair or a RGB-RGB pair, can significantly increase the person Re-ID performance, which validates its efficacy. Particularly, the effectiveness of each module in our framework has been validated in the ablation studies of our paper.
Second, the proposed ``RGB-RGB pair'' in Table~\ref{tab-8} means that the
two input branches of our framework are both set as RGB images. That is, we purposely replace the greyscale image in the upper branch with its original RGB image to demonstrate the advantage of using the former. We observe that ``RGB-Grey pair'' consistently outperforms ``RGB-RGB pair'' on all the benchmark datasets. Particularly, the most obvious gain is achieved on the dataset MSMT17. As shown in Fig.~\ref{fig12}, MSMT17 is much larger than the rest three datasets in terms of the number of both cameras and images, and is therefore closer to real scenarios. In this sense, although the improvement of ``RGB-Grey pair'' over ``RGB-RGB pair'' is not that significant on these smaller-scale datasets, its significant improvement on MSMT17 clearly shows the superiority and potential for more challenging person Re-ID settings.
\\

\setlength{\tabcolsep}{6pt}
\begin{table}[htbp]
  \centering
  \caption{Evaluation of different inputs in the proposed framework on Market1501, DukeMTMC-reID (Duke), CUHK03-NP-Detect (CHUK03) and MSMT17, respectively. The best performance is \textbf{bold}.}
    \begin{tabular}{|c|cc|cc|cc|cc|}
    \toprule
    \midrule
    \multirow{2}[1]{*}{Input} & \multicolumn{2}{c|}{Market1501} & \multicolumn{2}{c|}{Duke} & \multicolumn{2}{c|}{CUHK03} & \multicolumn{2}{c|}{MSMT17} \\
\cmidrule{2-9}          & mAP   & Rank-1 & mAP   & Rank-1 & mAP   & Rank-1 & mAP   & Rank-1 \\
    \midrule
    Baseline & 78.5  & 90.6  & 68.5  & 82.1 & 61.0  & 62.9  & 44.3  & 69.3 \\
    \midrule
    RGB-RGB pair & 85.1  & 94.0  & 75.7  & 87.5  & 69.6  & 72.4  & 53.6  & 77.3 \\
    \midrule
    RGB-grey pair & \textbf{85.6} & \textbf{94.5} & \textbf{76.5} & \textbf{88.0} & \textbf{69.9} & \textbf{73.3} & \textbf{55.0} & \textbf{78.6} \\
    \bottomrule
    \end{tabular}%
  \label{tab-8}%
\end{table}%

\textbf{Comparison of different image types.}
We also replace the greyscale images by the HSV images.  As seen in Table~\ref{tab01}, using the HSV images cannot bring improvement when compared with the greyscale images because an HSV image still contains both color and greyscale information as a RBG image does. As a result, it will be affected in a smilar manner as an RBG image in terms of the inaccuracy of color information in discriminating some person images. The proposed framework only retains greyscale information and combines it with a RGB image, which can better mitigate the above issue. This has been well verified by the experimental result in Table~\ref{tab01}.
\\
\begin{table}[htbp]
  \centering
  \caption{Comparison between different image types on four datasets.}
    \begin{tabular}{|c|cc|cc|cc|cc|}
    \toprule
    \midrule
    \multirow{2}[1]{*}{Image Type} & \multicolumn{2}{c|}{Market1501} & \multicolumn{2}{c|}{DukeMTMC-reID} & \multicolumn{2}{c|}{CUHK03-Detect} & \multicolumn{2}{c|}{MSMT17}\\
\cmidrule{2-9}          & mAP   & Rank-1 & mAP   & Rank-1 & mAP   & Rank-1 & mAP   & Rank-1\\
    \midrule
    Grey+RGB  & \textbf{85.6} & \textbf{94.5} & \textbf{76.5} & \textbf{88.0} & \textbf{69.9} & \textbf{73.3} & \textbf{55.0} & \textbf{78.6} \\
    \midrule
    HSV+RGB   & 83.6  & 93.5  & 73.9  & 86.2 & 63.6  & 67.1  & 49.5  & 72.6\\
    \bottomrule
    \end{tabular}%
  \label{tab01}%
\end{table}%

\textbf{Evaluation of different color to greyscale conversion methods.}
The weight setting in our paper considers the different human perception/sensibility towards these three colors, which is also used in a popular python toolbox ``Pillow''\footnote{https://pillow.readthedocs.io/en/stable/}. Following the way of generating greyscale images by ``Pillow'', we give the weight (i.e., R -> 0.299; G -> 0.587 and B -> 0.114) in our paper. In addition, we also utilize the arithmetical average of R, G, and B to convert RGB images to greyscale images and employ them in our experiments as reported in Table~\ref{tab05}. Overall, the average weight setting is slightly poor when compared with our method.

\begin{table}[htbp]
  \centering
  \caption{Comparison of different color to greyscale conversion methods on four datasets. Note that ``Grey-avg'' denotes the average weight setting.}
    \begin{tabular}{|c|cc|cc|cc|cc|}
    \toprule
    \midrule
    \multirow{2}[1]{*}{Image Type} & \multicolumn{2}{c|}{Market1501} & \multicolumn{2}{c|}{DukeMTMC-reID} & \multicolumn{2}{c|}{CUHK03-Detect} & \multicolumn{2}{c|}{MSMT17} \\
\cmidrule{2-9}          & mAP   & Rank-1 & mAP   & Rank-1 & mAP   & Rank-1 & mAP   & Rank-1\\
    \midrule
    Grey (ours) & 85.6  & 94.5  & 76.5  & 88.0 & 69.9  & 73.3  & 55.0  & 78.6\\
    \midrule
    Grey-avg & 85.8  & 94.3  & 76.0  & 87.8 & 69.6  & 72.4  & 54.8  & 78.0\\
    \bottomrule
    \end{tabular}%
  \label{tab05}%
\end{table}%

\textbf{The evaluation on different backbones and various dimensions of features.}
We also evaluate two ResNet-related backbones in our framework, as reported in Table~\ref{tab02}. According to these results, we observe that using a deeper network can enhance the generalization ability of our method. Please note that to have a fair comparison to the state-of-the-art methods, which mostly utilize ResNet-50 as the backbone, we report all experimental results based on ResNet-50 in our paper. 
In addition, we also assess our method with various-dimensional features and report experimental results in Table~\ref{tab03}. As seen, the low-dimensional feature (e.g., 192) has poor performance. When the dimension is over 1024, the performance tends to be stable.

\begin{table}[htbp]
  \centering
  \caption{Evaluation on different backbones on four datasets.}
    \begin{tabular}{|c|cc|cc|cc|cc|}
    \toprule
    \midrule
    \multirow{2}[1]{*}{Backbone} & \multicolumn{2}{c|}{Market1501} & \multicolumn{2}{c|}{DukeMTMC-reID} & \multicolumn{2}{c|}{CUHK03-Detect} & \multicolumn{2}{c|}{MSMT17}\\
\cmidrule{2-9}          & mAP   & Rank-1 & mAP   & Rank-1 & mAP   & Rank-1 & mAP   & Rank-1\\
    \midrule
    Resnet-50 & 85.6  & 94.5  & 76.5  & 88.0  & 69.9  & 73.3  & 55.0  & 78.6\\
    \midrule
    Resnet-101 & \textbf{87.9} & \textbf{94.8} & \textbf{78.6} & \textbf{89.3} & \textbf{73.5} & \textbf{77.1} & \textbf{59.1} & \textbf{81.3}\\
    \bottomrule
    \end{tabular}%
  \label{tab02}%
\end{table}%

\begin{table}[htbp]
  \centering
  \caption{Evaluation on various feature dimensions on four datasets.  Note that ``RGB+Grey+Joint'' denotes the dimensions of the RGB branch, the greyscal branch and the joint branch in our framework, respectively.}
    \begin{tabular}{|c|cc|cc|}
    \toprule
    \midrule
    \multirow{2}[1]{*}{RGB+Grey+Joint} & \multicolumn{2}{c|}{Market1501} & \multicolumn{2}{c|}{DukeMTMC-reID} \\
\cmidrule{2-5}          & mAP   & Rank-1 & mAP   & Rank-1 \\
    \midrule
    64+64+64=192 & 81.1  & 91.3  & 71.1  & 84.4 \\
    128+128+256=512 & 85.0  & 93.8  & 75.1  & 87.3 \\
    256+256+512=1024 & 85.8  & \textbf{94.6}  & 76.0  & 87.5 \\
    256+512+512=1280 & \textbf{86.0}  & 94.5  & 75.8  & \textbf{88.1} \\
    512+256+512=1280 & 85.6  & 94.5  & \textbf{76.5}  & 88.0 \\
    512+512+512=1536 & 85.8  & 94.3  & 75.9  & 86.8 \\
    \midrule
    \midrule
    \multirow{2}[1]{*}{RGB+Grey+Joint} & \multicolumn{2}{c|}{CUHK03-Detect} & \multicolumn{2}{c|}{MSMT17} \\
\cmidrule{2-5}          & mAP   & Rank-1 & mAP   & Rank-1 \\
    \midrule
    64+64+64=192 & 65.3  & 67.9  & 47.3  & 71.9 \\
    128+128+256=512 & 70.0  & 72.9  & 53.4  & 77.3 \\
    256+256+512=1024 & 69.7  & 73.3  & \textbf{55.0}  & 78.4 \\
    256+512+512=1280 & \textbf{70.6}  & \textbf{74.8}  & 54.9  & \textbf{78.6} \\
    512+256+512=1280 & 69.9  & 73.3  & \textbf{55.0}  & \textbf{78.6} \\
    512+512+512=1536 & 69.8  & 73.6  & 54.5  & 78.4 \\
    \bottomrule
    \end{tabular}%
  \label{tab03}%
\end{table}%

\textbf{The evaluation on data augmentation with greyscale images.}
To investigate this issue, we specifically leverage the greyscale images in the way of data augmentation and obtain results in Table~\ref{tab04}. Note that ``RGB:Grey'' indicates the ratio between RGB and greyscale images in a training set. For example, for the ``1:0.05'', we randomly choose 5\% RGB images to convert them into the greyscale images and use all RGB images and these greyscale images to train the model together. From these experimental results, we can see that it is not effective to simply use the greyscale image via data augmentation due to the noisy information from the greyscale image. In other words, different from our proposed framework, the conventional framework cannot effectively explore the complementarity between greyscale images and RGB images via the data-augmentation scheme. Therefore, we develop a novel deep framework to handle the issue.\\

\begin{table}[htbp]
  \centering
  \caption{Results of the data augmentation based on greysacle images on four datasets. Note that ``RGB:Grey'' denotes the ratio between RGB and greyscale images in a training set.}
    \begin{tabular}{|c|cc|cc|cc|cc|}
    \toprule
    \midrule
    \multirow{2}[1]{*}{RGB: Grey} & \multicolumn{2}{c|}{Market1501} & \multicolumn{2}{c|}{DukeMTMC-reID} & \multicolumn{2}{c|}{CUHK03-Detect} & \multicolumn{2}{c|}{MSMT17}\\
\cmidrule{2-9}          & mAP   & Rank-1 & mAP   & Rank-1 & mAP   & Rank-1 & mAP   & Rank-1 \\
    \midrule
    1~:~0 & 78.5  & 90.6  & 68.5  & 82.1 & 61.0  & 62.9  & 44.3  & 69.3\\
    \midrule
    1~:~0.05 & 78.5  & 90.9  & 68.3  & 82.2 & 58.9  & 61.7  & 44.8  & 70.1\\
    1~:~0.10 & 77.7  & 90.8  & 67.9  & 82.3 & 58.1  & 60.9  & 44.9  & 70.4\\
    1~:~0.25 & 75.6  & 90.2  & 64.2  & 80.6 & 55.7  & 59.7  & 43.8  & 69.8\\
    1~:~1 & 68.7  & 86.6  & 54.7  & 73.8 & 47.8  & 51.3  & 37.0  & 64.7\\
    \bottomrule
    \end{tabular}%
  \label{tab04}%
\end{table}%

\textbf{The evaluation of different fusion schemes.}
As discussed in Section~\ref{Discussion}, we also investigate different tensor fusion schemes in our framework, which include element-wise plus, element-wise multiply, concatenated operation and the adaptive weighting method.  Particularly, for the concatenated operation, we also employ the SE-block~\cite{DBLP:conf/cvpr/HuSS18} to compress the channels. The experimental results are reported in Table~\ref{tab-6}. In addition, to implement the adaptive weighting method, we embed a weighted network into the proposed framework, which consists of two FC layers that are simultaneously trained with the proposed framework. Given a RGB tensor $T_{rgb}\in \mathbb{R}^{2048\times 24 \times 8}$ or a greyscale tensor $T_{grey}\in \mathbb{R}^{2048\times 24 \times 8}$, the input is the 2048-d feature from GAP of $T_{rgb}$ or $T_{grey}$, and output is the weight value for greyscale or RGB tensors, respectively. We employ softmax to normalize them (i.e., $w_{rgb}$ and $w_{grey}$), and generate the joint tensor $T_{joint} = w_{grey}\times T_{grey} + w_{rgb}\times T_{rgb}$. This method is implemented by ourselves. As seen in Table~\ref{tab-6}, all results do not achieve significantly better performance than the proposed one. This is because the joint learning framework, using the independent loss in each branch and a global loss for the concatenated feature, has guaranteed the proposed network can obtain the robust features for person Re-ID, as discussed in Section~\ref{Discussion}.

\setlength{\tabcolsep}{12pt}
\begin{table}[htbp]
  \centering
  \caption{Evaluation of different fusion schemes on Market1501 and DukeMTMC-reID, respectively.}
    \begin{tabular}{|c|cc|cc|}
    \toprule
    \midrule
    \multirow{2}[2]{*}{Methods} & \multicolumn{2}{c|}{~~~Market1501~~} & \multicolumn{2}{c|}{~~DukeMTMC-reID~~} \\
\cmidrule{2-5}          & mAP   & Rank-1 & mAP   & Rank-1 \\
    \midrule
    Concatenate & 85.5  & 94.2  & 76.2  & 87.9 \\
    Concatenate+SE & 85.5  & 94.1  & 75.5  & 87.1 \\
    Multiply   & \textbf{85.8} & 94.3  & 76.4  & \textbf{88.2} \\
    Adaptive weight   & 85.6 & \textbf{94.5}  & 75.7  & 87.1 \\
    Plus (ours) & 85.6  & \textbf{94.5} & \textbf{76.5} & 88.0 \\
    \bottomrule
    \end{tabular}%
  \label{tab-6}%
\end{table}%


\textbf{The visualization of the RGB and greyscale branches.}
\begin{figure}
\centering
\subfigure[CUHK03]{
\includegraphics[width=6.5cm]{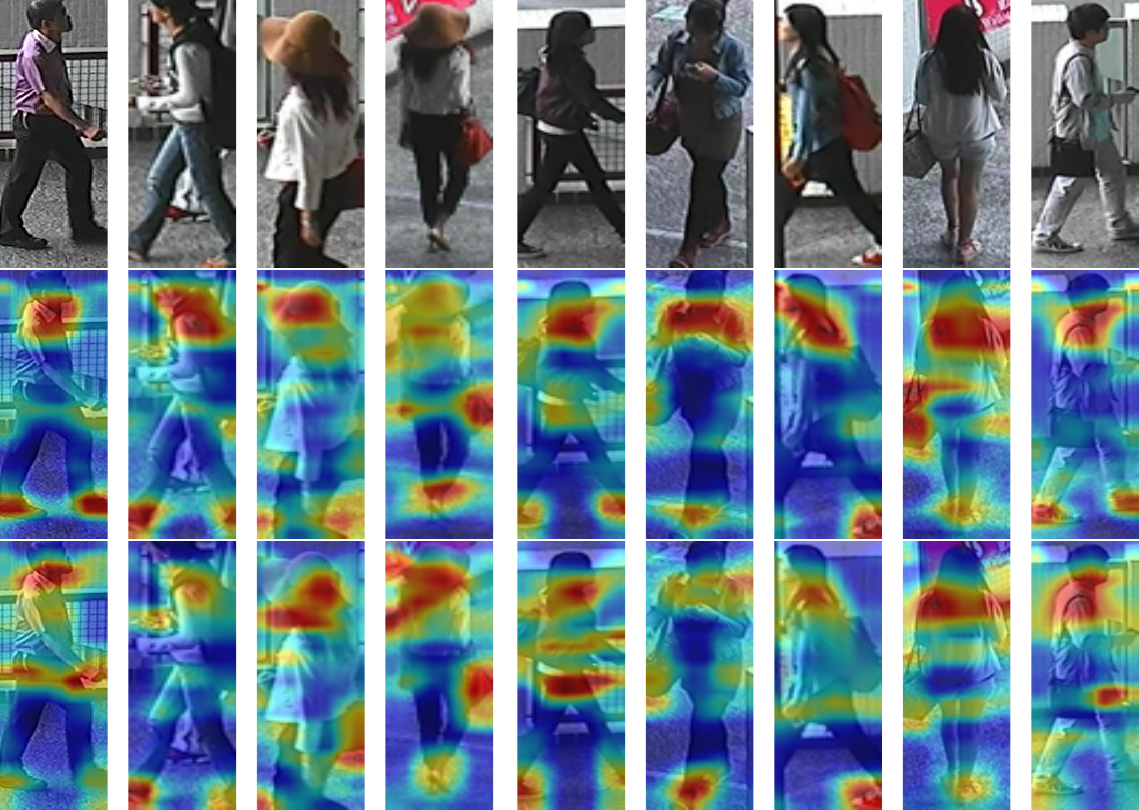}
}
\subfigure[DukeMTMC-reID]{
\includegraphics[width=6.5cm]{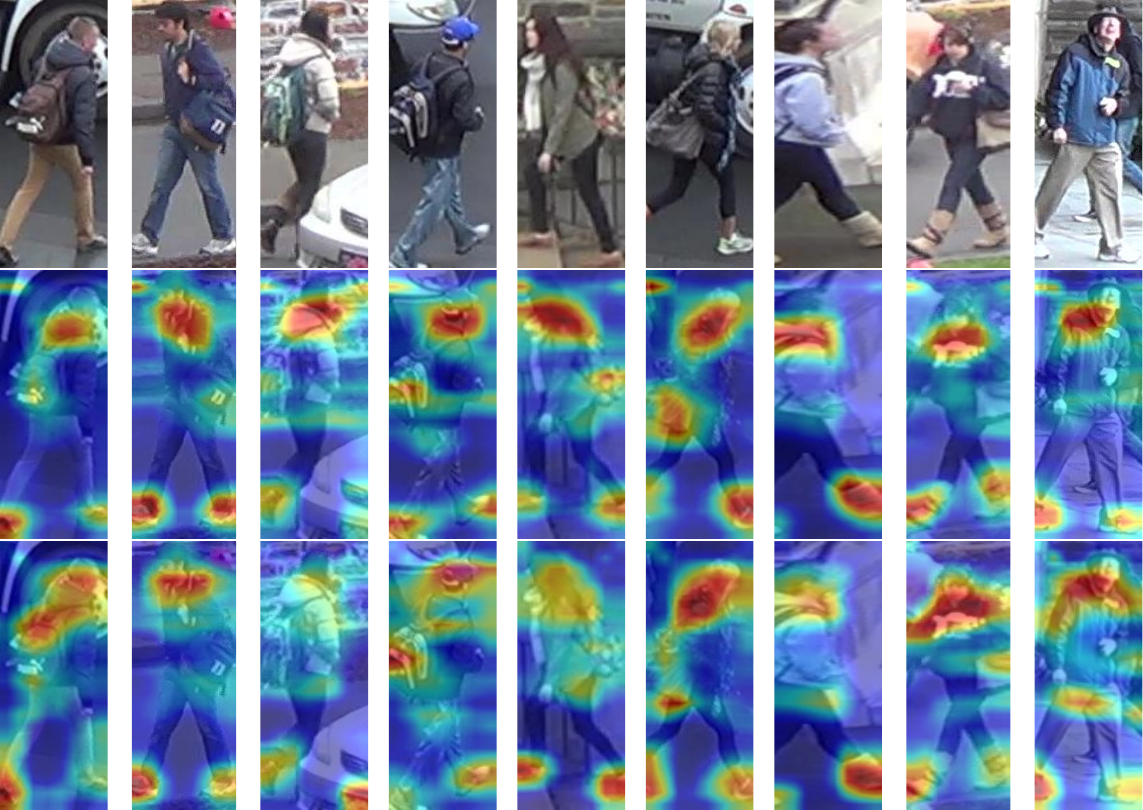}
}
\caption{Feature response maps of the last convolutional layer in the RGB and greyscale branches of the proposed method, respectively. Note that these images are from the testing set of CUHK03 and DukeMTMC-reID. Top denotes RGB images.  Middle and bottom are feature response maps of the greyscale and RGB branches, respectively. Note that the complementarity between them.}
\label{fig6}
\end{figure}
To further validate the complementarity between RGB and greyscale images, we visualize feature maps of the last convolutional layer in the greyscale and RGB branches of the proposed method, as shown in Fig.~\ref{fig6}. Since the conventional methods mainly focus on the color information, most false-positive samples are caused by the color similarity among many different person images, as shown in Fig.~\ref{fig1}. By removing the color information, the greyscale branch can enforce the network to pay more attention to other information besides color. For example, in Fig.~\ref{fig6}, we can observe that the greyscale and RGB branches of the proposed method focus on different regions. The complementarity further demonstrates that greyscale is great in the person Re-ID task.

\section{Conclusion}\label{s-conclusion}
In this paper, we point out that there is a complementarity between RGB and greyscale images in person Re-ID. To fully exploit the information in greyscale and RGB images, we propose a two-stream network with RGB-grey information. It can effectively combine the color and structure information to produce a robust representation for person Re-ID. The extensive experiments demonstrate not only the superiority of the proposed framework, but also the complementarity between RGB and greyscale images in person Re-ID on four large-scale benchmark datasets. It is hoped that this work can inspire more work on fully exploiting the greyscale information to help feature representation learning in person Re-ID.

{\small
\bibliographystyle{acm}
\bibliography{egbib}
}
\end{document}